# ACZ-GSeg: Adaptive Concentric Zone-based Two-stage Ground Segmentation for LiDAR Point Clouds

Ge Zhang[1], Chunyang Wang[2*], Bin Liu[1]

[1] School of Optoelectronic Engineering, Xi'an University of Technology, Shaanxi, Xi'an, 710000;

[2] Xi'an University of Technology, Xi'an Key Laboratory of Active Optoelectronic Imaging Detection Technology, Shaanxi, Xi'an, 710021;

**Abstract** Ground segmentation is a fundamental prerequisite for autonomous navigation, environmental perception, and object detection in ground mobile platforms. To address the under-segmentation of ground points caused by sparse long-range point clouds, ground undulations, and interference from non-ground structures in complex road scenarios, this paper proposes a two-stage ground segmentation method based on the Adaptive Concentric Zone Model. First, an Adaptive Concentric Zone Model is constructed to dynamically determine the number of sectors in each ring, thereby forming local zones with more balanced point distributions. Based on this model, a two-stage ground segmentation method is developed. In the coarse segmentation stage, a lowest-height seed constraint and height-decay weighting are introduced to establish a weighted principal component analysis plane fitting model, from which ground candidate points are extracted. In the fine segmentation stage, a reflectance intensity consistency constraint is employed to distinguish high-confidence ground points from uncertain points, and the uncertain points are further refined based on the local height stability of high-confidence neighborhoods. Experimental results show that the proposed method achieves Precision, Recall, and F1-score values of 99.12%, 96.24%, and 97.66% on the SemanticKITTI dataset, and 98.72%, 100.00%, and 99.36%, respectively, on a self-collected point cloud acquired using a RUBY-PLUS. The results demonstrate that the proposed method can effectively adapt to the range-dependent distribution characteristics of LiDAR point clouds, which are dense at near ranges and sparse at far ranges. It reduces the misclassification of non-ground points while maintaining ground point recall, thereby effectively improving the stability of ground segmentation.

# 1 Introduction

In recent years, with the rapid development of intelligent mobile platforms such as unmanned ground vehicles, unmanned aerial vehicles, mobile robots, and autonomous vehicles, the demand for stable and accurate three-dimensional perception of surrounding environments has continuously increased. To achieve environmental understanding and autonomous decision-making in complex scenarios, 3D perception methods have been widely applied to semantic segmentation, object detection, object tracking, and scene mapping [1],[2],[3],[4]. Among them, Light Detection and Ranging (LiDAR) can actively acquire high-precision 3D spatial information and offers several advantages, including high ranging accuracy, a wide field of view, robustness to illumination variations, and strong long-range sensing capability. It has therefore become an important sensor in environmental perception systems for mobile platforms [5][6]. Based on 3D point clouds acquired by LiDAR, extensive studies have been conducted on semantic segmentation [7][8], object tracking [9], and 3D object detection [10].

In 3D point cloud perception tasks, ground segmentation is an important fundamental preprocessing step [11][12]. On the one hand, accurate ground segmentation can provide traversable area estimation for mobile platforms, thereby supporting path planning and autonomous navigation [13]. On the other hand, ground segmentation can effectively assist non-ground object detection and tracking. Since traffic participants such as vehicles and pedestrians usually maintain spatial contact with the ground region [14], when ground points are accurately estimated, non-ground points can be further analyzed at the object level through clustering, detection, or tracking algorithms[15]. In addition, in outdoor road scenarios, a large proportion of point clouds are usually distributed on the ground surface. If ground points are effectively removed during preprocessing, the computational burden of subsequent object segmentation, detection, and tracking tasks can be significantly reduced [16]. Therefore, the ground region considered in this paper includes not only road surfaces but also local supporting regions that moving objects may contact, such as sidewalks, road shoulders, and grass areas.

With the development of deep learning, learning-based 3D point cloud perception methods have achieved favorable performance in ground segmentation and related tasks [17][18]. However, these methods still have several limitations. First, learning-based methods generally rely on large-scale point-wise annotated data, resulting in high dataset construction costs and labor-intensive annotation processes. Second, when there are differences between the test scenario and the training data in terms of road structure, sensor type, or point cloud distribution characteristics, model performance may degrade, indicating limited generalization capability across different mobile platforms and complex road environments. Therefore, it remains important to develop non-learning-based ground segmentation methods with strong scene adaptability, low data dependence, and good real-time performance.

As a prerequisite for subsequent 3D object detection, clustering, and tracking tasks, ground segmentation methods usually need to satisfy the following requirements. First, the algorithm should have high computational efficiency to meet the real-time requirements of online perception for mobile platforms. Second, the algorithm should maintain a good balance between precision and recall, avoiding excessive removal of true ground points while reducing the misclassification of non-ground structures as ground. Finally, the algorithm should be able to adapt to complex conditions in real road scenarios, such as ground undulations, slope variations, curb structures, and sparse long-range point clouds. However, existing non-learning-based ground segmentation methods are still prone to under-segmentation or mis-segmentation in complex scenarios. This is mainly because LiDAR point clouds exhibit a pronounced range-dependent density distribution, with dense points at near ranges and sparse points at far ranges. Fixed-scale local partitions are difficult to adapt simultaneously to both near-range dense regions and far-range sparse regions. In addition, non-ground structures such as curbs, vehicle underbodies, low obstacles, and vegetation may interfere with local plane estimation, causing true ground points to be incorrectly removed or non-ground points to be misclassified as ground.

To address the above issues, this paper proposes a two-stage ground segmentation method based on the Adaptive Concentric Zone Model. First, adaptive concentric rings are constructed according to the radial distance distribution of the point cloud, and the number of sectors in each ring is dynamically determined based on the principle of approximately constant arc length. In this way, local zones with relatively consistent scales and more balanced point distributions are formed. Subsequently, in the coarse segmentation stage, a lowest-height seed constraint and height-decay weighting are introduced into each local zone to establish a weighted principal component analysis plane fitting model. Ground candidate points are then obtained by combining the planarity constraint, normal angle constraint, and point-to-plane distance criterion. In the fine segmentation stage, a reflectance intensity consistency constraint is further used to distinguish high-confidence ground points from uncertain points. The uncertain points and misclassified high-elevation points are then refined based on the local height stability of high-confidence neighborhoods and a robust height threshold. Through these designs, the proposed method can better adapt to the range-dependent density distribution of LiDAR point clouds and reduce the interference of non-ground structures in local ground estimation, thereby improving the robustness and stability of ground segmentation in complex road scenarios.

The main contributions of this paper are summarized as follows:

(1) An Adaptive Concentric Zone Model is constructed. To address the uneven spatial distribution of LiDAR point clouds caused by range-dependent density variations, concentric rings are adaptively divided based on radial distance quantiles, and the number of sectors in different rings is dynamically determined according to the principle of approximately constant arc length. This forms local zones with relatively consistent scales and more balanced point distributions, thereby improving the stability of local ground structure modeling in complex road scenarios.

(2) A two-stage ground segmentation method is proposed. In the coarse segmentation stage, a lowest-height seed constraint and height-decay weighting are introduced to construct a weighted principal component analysis plane fitting model. Ground candidate points are obtained by combining the planarity constraint, normal angle constraint, and point-to-plane distance criterion. In the fine segmentation stage, a reflectance intensity consistency constraint is further used to distinguish high-confidence ground points from uncertain points. The uncertain points and misclassified high-elevation points are refined based on the local height stability of high-confidence neighborhoods and a robust height threshold, thereby improving the completeness and reliability of ground segmentation results.

(3) The effectiveness of the proposed method is validated on the public SemanticKITTI dataset and a self-collected point cloud acquired using a laboratory RUBY PLUS LiDAR. Experimental results show that the proposed method achieves Precision, Recall, and F1-score values of 99.12%, 96.24%, and 97.66%, respectively, on the SemanticKITTI dataset, and 98.72%, 100.00%, and 99.36%, respectively, on the self-collected RUBY PLUS point cloud. These results demonstrate the effectiveness of the proposed method on both public benchmark data and real road point clouds, indicating favorable segmentation accuracy, scene adaptability, and robustness.

# 2 Related Work

Most existing ground segmentation methods rely on geometric structural assumptions, elevation statistical features, or local neighborhood relationships for ground discrimination. These methods can be broadly categorized into geometry model-based methods, elevation grid-based methods, region- and graph-based modeling methods, and clustering- and multi-feature fusion-based methods.

Geometry model-based methods usually estimate ground structures based on local planes, scan lines, or terrain continuity assumptions. RANSAC [19] iteratively fits candidate planes through random sampling and regards the largest set of inliers satisfying a distance threshold as the ground region, showing good robustness in flat road scenarios. However, this method generally relies on a single-plane assumption, making it difficult to accurately describe ramps, undulating roads, and local uneven regions. In addition, the random iterative process introduces a certain computational overhead. Moosmann et al. [20] segmented 3D point clouds based on local convexity constraints, enabling rapid ground recognition in non-flat urban environments. Himmelsbach et al. [21] adopted a scan-line-based ground segmentation strategy and improved segmentation efficiency by analyzing slope variations between adjacent points. Zermas et al. [22] proposed a fast ground plane fitting method, namely GPF, which initializes the ground region using lowest-height seed points and estimates ground plane parameters through principal component analysis, thereby avoiding extensive random iterations. JCP [23] projects LiDAR points onto a 2D range view and applies a slope-based heuristic method to obtain an elevation variation map as the input for coarse-to-fine ground segmentation. Narksri et al. [24] further adopted a multi-region plane fitting strategy to adapt to sloped terrain, while Jimenez et al. [25] introduced a channel-based Markov random field to model local geometric variations. Although these methods improve the adaptability of traditional plane models to complex terrains, plane estimation remains susceptible to interference when local regions simultaneously contain non-ground structures such as curbs, vehicle underbodies, low obstacles, or vegetation. This may further lead to the under-segmentation of ground points or the misclassification of non-ground points as ground.

Elevation grid-based methods project 3D point clouds onto 2D grids or height maps and perform ground classification based on intra-cell height differences, local height variance, or terrain continuity. GroundGrid [26] maps point clouds into an elevation grid space, detects ground cells through local height statistics, and fills missing regions using an interpolation strategy, offering high computational efficiency and convenient engineering implementation. B-TMS [27] models terrain height using a Bayesian generalized kernel function on a Tri-Grid Field, while GATA [28] adopts a probabilistic graphical model for real-time terrain estimation. Such methods can transform unstructured point clouds into regularized representations, facilitating efficient processing. However, their performance is usually sensitive to grid resolution and height thresholds. When a single grid cell contains both ground and non-ground points, cell-level statistical features can be affected by mixed structures. In far-range sparse regions or areas with significant local undulations, the terrain continuity assumption of height maps may also become invalid.

Region- and graph-based modeling methods enhance the local adaptability of ground estimation by constructing local subregions or adjacency graphs. PatchWork [29] adopts a concentric zone model to divide LiDAR scans into multiple polar regions, performs principal component analysis-based ground plane fitting within each region, and then classifies ground and non-ground points using a probabilistic likelihood test. PatchWork++ [30] further introduces adaptive parameter adjustment, temporal consistency constraints, and enhanced strategies for noise and multi-layer ground, improving segmentation robustness in complex scenarios. TRAVEL [31] represents point clouds as node-edge relationships using a Tri-Grid Field graph structure and models traversable ground by analyzing concave-convex relationships between adjacent nodes and line segments. Compared with global plane fitting methods, regionalized or graph-structured methods can better characterize local terrain variations. However, their segmentation performance still depends on region scale, graph connectivity, and local point cloud density. When ground and non-ground structures coexist within a partition, or when far-range point clouds are excessively sparse, the local region may be misclassified as a whole, thereby affecting segmentation completeness.

Clustering- and multi-feature fusion-based methods perform ground discrimination from the perspectives of point cloud density, local geometric consistency, or multi-source feature representation. DBSCAN [32] achieves unsupervised clustering based on density-reachable relationships and can be used to separate locally continuous ground regions from discrete non-ground structures. However, vehicle-mounted LiDAR point clouds exhibit a pronounced range-dependent density distribution. Fixed neighborhood radii and minimum point thresholds are difficult to adapt simultaneously to regions at different ranges, which may cause missed segmentation of far-range ground points or misclassification of near-range non-ground points. FusSeg [33] enhances ground recognition by fusing features such as height, distance, normals, intensity, and local geometric consistency. Compared with single geometric constraints, it provides richer discriminative information. Nevertheless, its performance is still affected by feature weights, threshold settings, and differences in sensor intensity calibration, which may lead to insufficient parameter adaptability in cross-scenario applications.

In summary, existing methods have certain advantages in real-time performance, interpretability, and engineering deployment, but they still face three key challenges in complex road scenarios. First, fixed-scale partitions or fixed density thresholds are difficult to adapt to the range-dependent spatial distribution of vehicle-mounted LiDAR point clouds, which are dense at near ranges and sparse at far ranges. Second, non-ground structures in locally mixed regions can interfere with ground model estimation, resulting in unstable segmentation boundaries. Third, single geometric or elevation constraints are difficult to balance ground point recall and non-ground point suppression simultaneously. To address these issues, this paper proposes a two-stage ground segmentation method based on the Adaptive Concentric Zone Model. The proposed method improves the modeling balance across different distance regions through the Adaptive Concentric Zone Model and enhances the discrimination capability for uncertain regions by combining coarse segmentation and fine segmentation mechanisms, thereby improving the robustness and stability of ground segmentation in complex road scenarios.

# 3 Methodology of ACZ-GSeg

The proposed method mainly consists of three steps, and its overall framework is shown in Fig. 1. First, in the Adaptive Concentric Zone stage, the original point cloud is projected onto the horizontal plane. An adaptive ring partitioning strategy is then constructed based on radial distance quantiles, and the number of sectors in each ring is dynamically determined according to the principle of approximately constant arc length, thereby forming local zones with relatively consistent scales and more balanced point distributions. In the coarse segmentation stage, a lowest-height seed constraint and height-decay weighting are introduced for each local zone to establish a weighted principal component analysis plane fitting model. Ground candidate points are then obtained by combining the planarity constraint, normal angle constraint, and point-to-plane distance criterion. In the fine segmentation stage, a reflectance intensity consistency constraint is used to divide the ground candidate points into high-confidence ground points and uncertain points, and the uncertain points are further discriminated based on the local height stability of high-confidence neighborhoods.

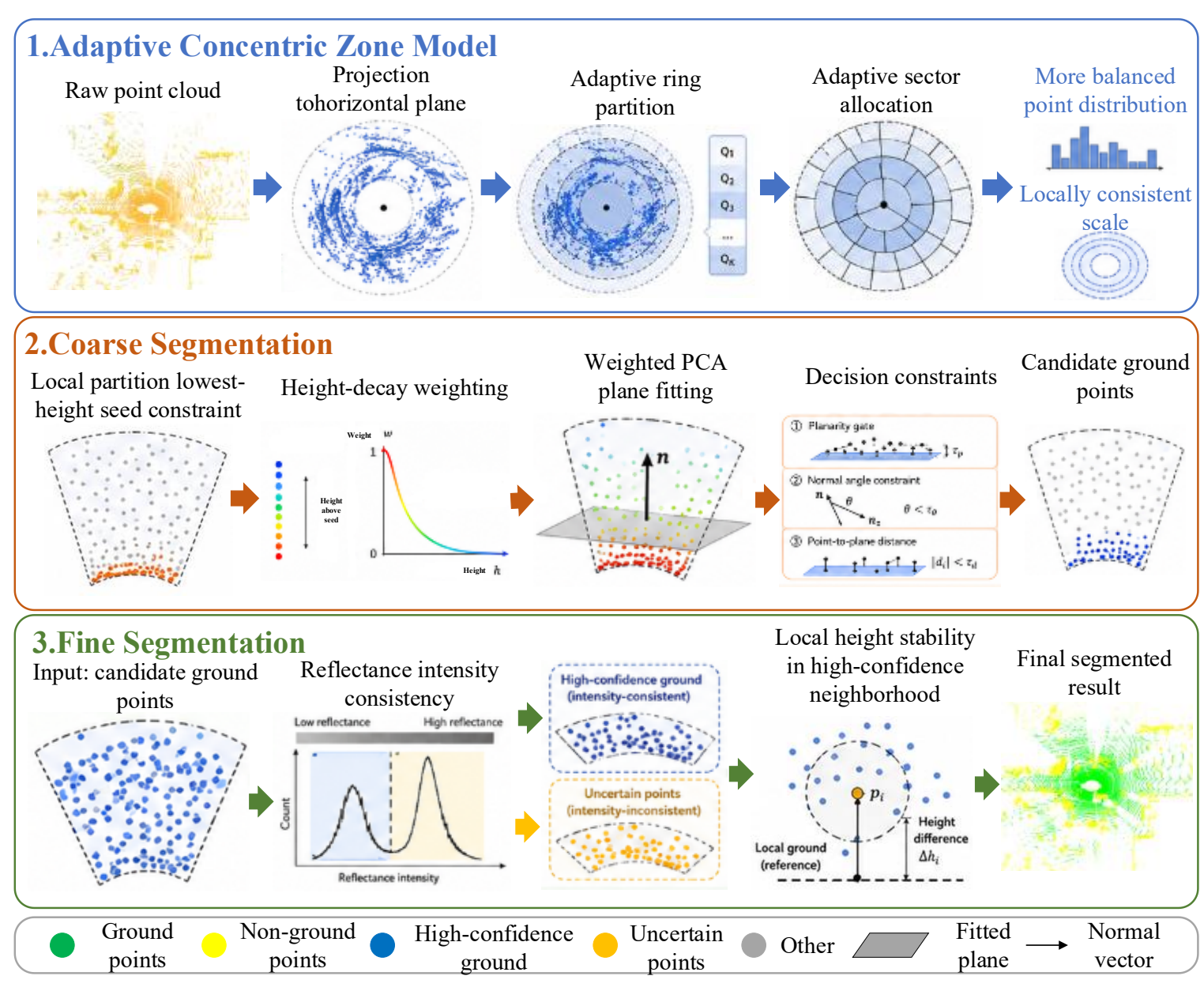

Fig. 1 Overall framework of ACZ-GSeg

## 3.1 Adaptive Concentric Zone Model

LiDAR point clouds usually exhibit a pronounced range-dependent density distribution, where points are densely distributed in near-range regions but become significantly sparser in far-range regions due to angular resolution, sensing distance, and target reflectance characteristics. If fixed-width rings or fixed-angle sectors are directly used for local partitioning, near-range zones may contain excessive points, resulting in overly complex local structural representations, whereas far-range zones may suffer from insufficient points or even empty partitions, thereby reducing the stability of subsequent local plane fitting. To address this issue, this paper constructs an Adaptive Concentric Zone Model. Specifically, rings are divided according to radial distance quantiles, and the number of sectors in each ring is dynamically determined based on the principle of approximately constant arc length. In this way, a partition structure with more balanced point distributions and relatively consistent local scales is established.

Fig. 2 compares the structural differences between the Concentric Zone Model and the proposed Adaptive Concentric Zone Model. The Concentric Zone Model, represented by the CZM adopted in Patchwork and Patchwork++, usually divides the radial space into four empirically predefined zones, with preset ring widths and sector scales assigned to different zones. This strategy is structurally simple and facilitates efficient regional processing. However, its zone boundaries and partition scales mainly rely on empirical settings and are difficult to adjust dynamically according to the spatial distribution characteristics of the input point cloud. For vehicle-mounted LiDAR point clouds with non-uniform density distributions, i.e., dense near-range regions and sparse far-range regions, fixed partitions can easily lead to point redundancy in near-range regions and insufficient samples in far-range regions, thus affecting the stability of local ground plane modeling. In contrast, the proposed Adaptive Concentric Zone Model no longer uses fixed empirical boundaries. Instead, it adaptively determines ring boundaries according to the radial distance distribution of the input point cloud, making the number of points within each ring more balanced. Meanwhile, the number of sectors in different rings is dynamically determined according to a target arc-length constraint, so that the angular partition scale can be adaptively adjusted with the radius. The resulting Adaptive Concentric Zone Model better matches the spatial density variation of real point clouds, improves the consistency and reliability of local ground modeling in both near- and far-range regions, and provides a more stable local partitioning basis for subsequent weighted PCA plane fitting and two-stage ground segmentation.

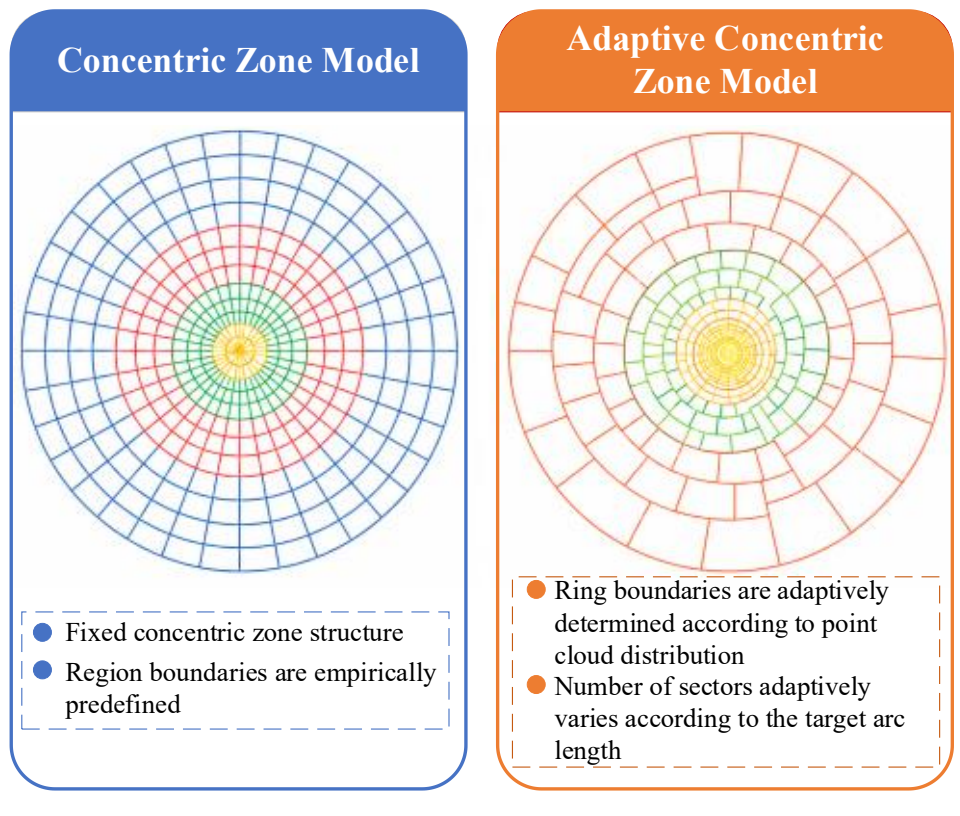

Fig. 2 Comparison between the Concentric Zone Model and the Adaptive Concentric Zone Model.

Let a frame of point cloud be denoted as $P=\{p_i\}_{i=1}^{N}$, $p_i=[x_i,y_i,z_i,I_i]^{\mathrm{T}}\in\mathbb{R}^4$ ,where $N$ is the number of points, $(x_i,y_i,z_i)$ denotes the 3D coordinates of point $p_i$, and $I_i\in R$ represents its reflectance intensity. If reflectance intensity is unavailable, $I_i$ is set to 0 and is excluded from the subsequent intensity constraint stage. The objective of ground segmentation is to divide the point cloud into a ground point set $G$ and a non-ground point set $N$.

$$P = G \bigcup N, G \bigcap N = \varnothing \tag{1}$$

**3.1.1 Horizontal Plane Projection and Polar Coordinate Representation**

To characterize the radial distribution of the point cloud on the horizontal plane, the 3D point cloud is first projected onto the $xOy$ plane, and the polar coordinate representation of each point is computed as follows:

$$r_i = \sqrt{x_i^2 + y_i^2} \tag{2}$$

$$\theta_i = a\tan 2\left(y_i, x_i\right) \in \left[-\pi, \pi\right] \tag{3}$$

where $r_i$ denotes the horizontal radial distance from the $i$-th point to the LiDAR coordinate origin, and $\theta_i$ represents the azimuth angle of this point on the horizontal plane. Compared with directly partitioning the 3D space using a regular grid, the polar coordinate representation on the horizontal plane can more naturally describe the scanning distribution characteristics of rotating LiDAR point clouds, thereby providing a coordinate basis for subsequent ring and sector partitioning.

**3.1.2 Adaptive Ring Partitioning Based on Radial Distance Quantiles**

The Concentric Zone Model usually adopts fixed radial intervals, in which the boundaries of adjacent rings satisfy:

$$r_k - r_{k-1} = \Delta r \tag{4}$$

where $\Delta r$ denotes the fixed ring width. Since the density of LiDAR point clouds decreases with increasing distance, fixed ring widths may result in an excessive number of points in near-range rings but insufficient points in far-range rings, leading to a severely imbalanced number of samples for local modeling across different rings. The stability of local plane fitting is closely related to the number of points involved in the fitting process. When the number of points in far-range rings is insufficient, the estimated plane normal and distance parameters are more susceptible to noise points or local outliers.

To address the above issue, this paper adopts an adaptive ring partitioning strategy based on radial distance quantiles. Let the desired number of rings be $K_0$. To avoid generating a large number of invalid rings with too few points in far-range regions, the effective number of rings is determined according to the total number of points $N$ and the minimum expected number of points per ring $N_{\min}$:

$$K = \min\left(K_0, \max\left(K_{\min}, \lfloor N / N_{\min} \rfloor\right)\right) \tag{5}$$

where $K$ denotes the actual number of rings used for the current point cloud frame, $K_0$ is the preset maximum number of rings, $K_{\min}$ is the minimum number of rings used to ensure the basic radial representation capability of the partition, $N_{\min}$ is the minimum expected number of points in each ring, and $\lfloor \cdot \rfloor$ denotes the floor operation. This design adaptively adjusts the number of rings according to the scale of the current point cloud frame, thereby avoiding excessive empty rings or weakly constrained rings in sparse point cloud frames.

Let the set of radial distances of all points be denoted as:

$$D_r = \left\{r_i\right\}_{i=1}^{N} \tag{6}$$

The quantile sequence is defined as follows:

$$q_k = \frac{k}{K}, k = 0,1,\ldots,K \tag{7}$$

Using the quantile function $Q_{D_r}\left(\bullet\right)$ of the radial distance sample set $D_r$, the ring boundaries are obtained as follows:

$$r^{(k)} = Q_{D_r}\left(q_k\right), k = 0,1,\ldots,K \tag{8}$$

The $k$-th ring is defined as:

$$R^{(k)} = \left\{i \middle| r_i \in \left[r^{(k-1)}, r^{(k)}\right]\right\}, k = 1,2,\ldots,K \tag{9}$$

where $r_k$ denotes the boundary of the $k$-th ring, and $Q_r\left(\bullet\right)$ represents the distance value corresponding to the quantile $q_k$ in the radial distance set $D_r$. Accordingly, the $k$-th adaptive ring is defined as

$$R_k = \left\{p_i \in P \middle| r^{(k-1)} \le r_i \le r^{(k)}\right\}, k = 1,2,\ldots,K-1 \tag{10}$$

For the outermost ring, to ensure that the points with the maximum radial distance are included, it is defined as

$$R_K = \left\{p_i \in P \middle| r^{(K-1)} \le r_i \le r^{(K)}\right\} \tag{11}$$

According to the definition of quantile-based partitioning,

when there are no large numbers of repeated radial distance values in the point cloud, the number of points in each ring approximately satisfies

$$\left|R_1\right| \approx \left|R_1\right| \approx \ldots \approx \left|R_K\right| \approx \frac{N}{K} \tag{12}$$

where $\left|R_K\right|$ denotes the number of points in the $k$ -th ring. Compared with fixed radial-width partitioning, quantile-based ring partitioning can automatically adjust the ring width according to the actual distribution of the point cloud. Specifically, in near-range regions with dense points, the ring width is relatively small, whereas in far-range regions with sparse points, the ring width increases accordingly. This design can avoid excessive aggregation in near-range regions and reduce invalid fitting in far-range regions caused by insufficient points, thereby improving the stability of subsequent local plane estimation.

In practical computation, some quantile boundaries may overlap due to repeated distance values or degenerated point cloud distributions, namely,

$$r_k = r_{k-1} \tag{13}$$

In this case, the corresponding ring cannot form a valid radial interval. Therefore, repeated boundaries can be merged, or the partitioning can degenerate into uniform radial partitioning when the number of valid boundaries is insufficient, so as to ensure the numerical stability of the partitioning process.

#### 3.1.3 Adaptive Sector Partitioning Based on Approximately Constant Arc Length

Radial ring partitioning alone cannot ensure that local zones have appropriate spatial scales. If a fixed number of sectors is used for all rings, the arc length corresponding to each sector increases with the radius, namely:

$$l_k = r_k \Delta\theta \tag{14}$$

where $l_k$ denotes the approximate arc length corresponding to a single sector in the k -th ring, $r_k$ denotes the representative radius of this ring, and $\Delta\theta$ represents the fixed angular interval. As indicated by Eq. (14), when $\Delta\theta$ is fixed, the transverse scale of sectors in far-range regions increases significantly. As a result, a single local zone may simultaneously contain ground points, curbs, vehicle underbodies, or other non-ground structures, thereby interfering with plane fitting.

To improve the consistency of local modeling scales across regions at different ranges, this paper adopts the principle of approximately constant arc length to adaptively assign the number of sectors to each ring. First, the mid-radius of the k-th ring is defined as:

$$\bar{r}_k = \frac{e_{k-1} + e_k}{2} \tag{15}$$

where $\bar{r}_k$ denotes the representative radius of the k -th ring.

Let the target arc length be $L_a$. Then, the number of sectors in the k -th ring is defined as:

$$M_k = clip\left(\left\lceil \frac{2\pi \max\left(\bar{r}_k, r_s\right)}{L_a} \right\rceil, M_{\min}, M_{\max}\right) \tag{16}$$

where $M_k$ denotes the number of sectors in the k -th ring, $L_a$ denotes the preset target arc length used to control the transverse scale of each local zone, and $r_s$ is the small-radius protection term used to prevent $\bar{r}_k$ in near-range regions from being too small and generating too few sectors. $M_{\min}$ and $M_{\max}$ denote the lower and upper bounds of the number of sectors, respectively. $\lceil \cdot \rceil$ represents the ceiling operation, and $clip(a,b,c)$ denotes truncating a to the interval $[b,c]$. Accordingly, the angular width of each sector in the k -th ring is given by:

$$\Delta\theta_k = \frac{2\pi}{M_k} \tag{17}$$

According to Eqs. (16) and (17), the approximate arc length of a single sector in the k-th ring is given by:

$$\bar{r}_k \frac{2\pi}{M_k} \approx L_a \tag{18}$$

Eqs. (18) indicates that, by allowing the number of sectors to adaptively vary with the ring radius, local zones at different ranges can maintain approximately consistent transverse scales. This design avoids the excessively large local support regions generated by fixed-angle partitioning in far-range areas, thereby reducing the probability that ground and non-ground structures are mixed within the same partition. Meanwhile, the upper bound $M_{\max}$ on the number of sectors prevents over-partitioning in far-range regions, ensuring that each local zone still contains sufficient points for subsequent plane estimation.

For any point $p_i \in R_k$, the index of the sector to which it

belongs is defined as:

$$m_i = \left\lfloor \frac{\theta_i + \pi}{2\pi} M_k \right\rfloor + 1, m_i \in \{1, ..., M_k\} \tag{19}$$

where $m_i$ denotes the sector index of point $p_i$ within the k-th ring. Since $\theta_i \in [-\pi, \pi)$, the angular range is shifted to $[0, \pi)$ by using $\theta_i + \pi$, which facilitates unified sector indexing. To avoid numerical boundary errors, when Eq. (19) gives $m_i = M_k + 1$, it is clipped to $M_k$.

Finally, the local zone corresponding to the k -th ring and the m -th sector is defined as:

$$B_{k,m} = \{p_i \in P | p_i \in R_k, m_i = m\}, k = 1, 2, ..., K, m = 1, 2, ..., M_k \tag{20}$$

where $B_{k,m}$ denotes the basic processing unit for local plane estimation in the subsequent coarse segmentation stage. For zones with fewer points than the minimum fitting point threshold $N_b^{\min}$, the corresponding zone can be skipped in the subsequent plane fitting stage or merged into a neighboring zone to avoid unstable estimation caused by insufficient samples.

In summary, the proposed Adaptive Concentric Zone simultaneously considers point-count balance and local-scale consistency. Ring partitioning based on radial distance quantiles alleviates the sample imbalance caused by the range-dependent density distribution of vehicle-mounted LiDAR point clouds. Sector partitioning based on the principle of approximately constant arc length controls the transverse scale of local zones at different ranges, thereby reducing the interference of mixed structures on local plane estimation. The resulting local zones provide a more stable, balanced, and physically meaningful local support region for the weighted PCA plane fitting in the subsequent coarse segmentation stage.

### 3.2 Coarse Segmentation Based on Height-Decay Weighted PCA for Ground Candidate Extraction

After Adaptive Concentric Zone partitioning, the original point cloud is divided into a series of local zones $B_{k,m}$, where $k = 1, 2, ..., K$ denotes the ring index, $m = 1, 2, ..., M_k$ denotes the sector index within the k -th ring, and $B_{k,m}$ represents the local point set corresponding to the m -th sector in the k -th ring. The objective of the coarse segmentation stage is to estimate a local ground plane within each local zone and extract ground candidate points based on geometric consistency constraints. Unlike global plane fitting performed directly on the entire point cloud, the ground surface within a local zone can usually be approximated by a local plane over a relatively small spatial range. Therefore, partition-wise plane fitting can better adapt to road slope variations, local undulations, and non-flat terrain.

#### 3.2.1 Lowest-Height Seed Constraint and Fitting Point Set Construction

For any local zone $B_{k,m}$, let the number of points be denoted as $N_{k,m} = |B_{k,m}|$ where $N_{k,m}$ represents the number of points contained in the local zone $B_{k,m}$. If $N_{k,m} < N_b^{\min}$, the number of points in this zone is insufficient to stably estimate a local plane. Therefore, this zone can be skipped or merged into a neighboring zone in practical implementation, where $N_b^{\min}$ denotes the minimum number of points required for local plane fitting.

For a local zone that satisfies the minimum point requirement, the minimum height within the zone is first computed as follows:

$$z_{k,m}^{\min} = \min_{p_i \in B_{k,m}} z_i \tag{21}$$

where $z_i$ denotes the vertical height coordinate of point $p_i$, and $z_{k,m}^{\min}$ denotes the minimum height within the local zone $B_{k,m}$. Since non-ground objects such as vehicles, pedestrians, and roadside facilities are usually located above the ground, whereas true ground points are often close to the minimum height within a local region, the minimum height can serve as an important reference for constructing the initial ground seed subset.

The height elevation of point $p_i$ relative to the minimum height of the local zone is further defined as:

$$\Delta z_i = z_i - z_{k,m}^{\min}, p_i \in B_{k,m} \tag{22}$$

where $\Delta z_i$ denotes the relative height of point $p_i$ above the minimum height of the corresponding local zone. Let the lowest-height seed band be $h_s$. Then, the initial seed point set used for local plane fitting is defined as:

$$S_{k,m} = \{p_i \in B_{k,m} | 0 \le \Delta z_i \le h_s\} \tag{23}$$

where $S_{k,m}$ denotes the seed point set close to the minimum height in the local zone $B_{k,m}$, and $h_s$ denotes the seed height band, which is used to control the range of low-height points involved in plane fitting.

The reason for introducing the lowest-height seed constraint is that, if all points within a local zone are directly used for plane fitting, non-ground structures such as vehicle underbodies, curbs, guardrails, and vegetation may significantly alter the principal direction of the covariance matrix, causing the estimated plane to deviate from the true ground surface. By preferentially selecting points near the minimum height for fitting, the dominant role of true ground points in plane estimation can be enhanced, while the interference of high-elevation structures on the local ground model can be reduced.

If $\left|S_{k,m}\right| < N_s^{\min}$, it indicates that the number of points satisfying the lowest-height constraint in this local zone is insufficient. To avoid unstable estimation caused by an insufficient number of samples, the fitting point set can be expanded to include the $N_s^{\min}$ points with the lowest heights in the local zone, or the corresponding zone can be skipped. Here, $N_s^{\min}$ denotes the minimum number of seed points required for weighted PCA plane fitting.

### 3.2.2 Height-Decay Weight Construction

Although the lowest-height seed set can suppress non-ground points that are significantly higher than the ground surface, points corresponding to curbs, vehicle underbodies, or low obstacles may still be included in the seed height band in complex road scenarios. To further reduce the influence of points with relatively large heights on plane estimation, height-decay weights are introduced in this paper. For any point $p_i$ in the fitting point set $S_{k,m}$, its weight is defined as:

$$\omega_i = \exp\left(-\alpha \Delta z_i\right), p_i \in S_{k,m} \tag{24}$$

where $\omega_i$ denotes the weight of point $p_i$ in local plane fitting, $\alpha > 0$ is the height-decay coefficient, and $\Delta z_i$ represents the height elevation of point $p_i$ relative to the local minimum height. According to Eq. (24), when a point is closer to the local minimum height, $\Delta z_i$ is smaller and its weight $\omega_i$ becomes larger. Conversely, as the relative height increases, its weight decays exponentially, thereby reducing its contribution to the estimation of plane parameters.

To provide a clear physical meaning for the weight parameter, the weight at the upper boundary $h_s$ of the seed height band can be set to $\eta$, namely:

$$\exp\left(-\alpha h_s\right) = \eta, 0 < \eta < 1 \tag{25}$$

Thus, the height-decay coefficient can be obtained as:

$$\alpha = -\frac{\ln \eta}{h_s} \tag{26}$$

where $\eta$ denotes the relative weight assigned to points located at the upper boundary of the seed height band with respect to the minimum-height point. Eq. (26) links the decay coefficient $\alpha$ with the seed height band $h_s$ and the boundary weight $\eta$, making the height-decay process interpretable. This design weakens the influence of relatively high points on the fitting result while preserving local ground structural information, thereby improving the robustness of local plane estimation against low obstacles and locally mixed structures.

### 3.2.3 Local Plane Fitting Based on Weighted PCA

After obtaining the fitting point set $S_{k,m}$ and its weights $\omega_i$, weighted principal component analysis is used to estimate the local ground plane. Let the 3D coordinate vector of point $p_i$ be denoted as:

$$\mathrm{X}_i = \left[x_i, y_i, z_i\right]^T \in \mathbb{R}^3 \tag{27}$$

The local ground plane is formulated as:

$$\Pi_{k,m} : n_{k,m}^T \left(\mathrm{X} - \mu_{k,m}\right) = 0 \tag{28}$$

where $\Pi_{k,m}$ denotes the local ground plane estimated in the local zone $B_{k,m}$, $n_{k,m} \in \mathbb{R}^3$ is the unit normal vector of this plane, satisfying $\left\|n_{k,m}\right\|_2 = 1$, $\mu_{k,m} \in \mathbb{R}^3$ represents a reference point on the plane, which is usually taken as the weighted centroid, and $\mathrm{X}$ denotes an arbitrary 3D point on the plane.

Weighted PCA-based plane fitting aims to minimize the weighted sum of squared orthogonal distances, namely:

$$\min_{n_{k,m}, \mu_{k,m}} J = \sum_{p_i \in S_{k,m}} \omega_i \left(n_{k,m}^T \left(\mathrm{X}_i - \mu_{k,m}\right)\right)^2, s.t. \left\|n_{k,m}\right\|_2 = 1 \tag{29}$$

where J denotes the objective function of weighted plane fitting, $n_{k,m}^T \left(\mathrm{X}_i - \mu_{k,m}\right)$ represents the signed orthogonal distance from point $p_i$ to the plane $\prod_{k,m}$, and the weight $\omega_i$ is used to control the contribution of different points to plane estimation. The physical meaning of Eq. (29) is that, under the unit-norm constraint on the normal vector, a local plane is sought such that the overall weighted orthogonal distance from the fitting points to the plane is minimized.

By optimizing the objective function with respect to $\mu_{k,m}$, the weighted centroid can be obtained as:

$$\mu_{k,m} = \frac{\sum_{p_i \in S_{k,m}} \omega_i \mathrm{X}_i}{\sum_{p_i \in S_{k,m}} \omega_i} \tag{30}$$

where $\mu_{k,m}$ denotes the weighted centroid of the fitting point set. Compared with the arithmetic mean, the weighted centroid is more biased toward low-height seed points, and therefore can better represent the spatial position of the local ground region.

The centered vector is further defined as:

$$\mathrm{X}_i = \mathrm{X}_i - \mu_{k,m} \tag{31}$$

where $\mathrm{X}_i$ denotes the offset vector of point $p_i$ with respect to the weighted centroid. Based on the centered vectors, the weighted covariance matrix can be constructed as:

$$C_{k,m} = \frac{\sum_{p_i \in S_{k,m}} \omega_i \mathrm{X}_i \mathrm{X}_u^T}{\sum_{p_i \in S_{k,m}} \omega_i} \tag{32}$$

where $C_{k,m} \in \mathbb{R}^{3\times3}$ denotes the weighted covariance matrix of the local fitting point set, which is used to describe the spatial distribution of points in different directions within the local zone. In the weighted covariance matrix, low-height points are assigned larger weights, and therefore have a stronger influence on the principal directions of the spatial distribution matrix.

By substituting Eq. (32) into Eq. (29), the plane fitting problem can be transformed into:

$$\min n_{k,m}^T c_{k,m} n_{k,m}, s.t. \|n_{k,m}\|_2 = 1 \tag{33}$$

According to the Rayleigh quotient minimization principle, the optimal solution of Eq. (33) is the eigenvector corresponding to the smallest eigenvalue of the weighted covariance matrix $C_{k,m}$. Let its eigenvalues be arranged in ascending order as:

$$\lambda_{k,m}^{(1)} \le \lambda_{k,m}^{(2)} \le \lambda_{k,m}^{(3)} \tag{34}$$

The corresponding unit eigenvectors are denoted as:

$$v_{k,m}^{(1)}, v_{k,m}^{(2)}, v_{k,m}^{(3)} \tag{35}$$

Thus, the local plane normal is given by:

$$n_{k,m} = v_{k,m}^{(1)} \tag{36}$$

where $\lambda_{k,m}^{(1)}$ denotes the minimum variance of the point cloud in the normal direction, and $v_{k,m}^{(1)}$ represents the direction corresponding to this minimum variance. For a local planar structure, the point cloud is mainly distributed along the two in-plane directions, while its variance along the normal direction is small. Therefore, the eigenvector corresponding to the smallest eigenvalue can be used as the normal vector of the local ground plane.

**3.2.4 Planarity Constraint**

In complex road scenarios, some local zones may contain vegetation, vehicle edges, roadside facilities, or other non-ground structures with irregular spatial distributions. Even with the lowest-height seed constraint and height-decay weighting, these zones may still fail to form a reliable local plane. Therefore, a planarity constraint is introduced to determine whether the current local zone has a sufficiently planar structure.

Based on the eigenvalues of the weighted covariance matrix, the local curvature or surface variation is defined as:

$$\kappa_{k,m} = \frac{\lambda_{k,m}^{(1)}}{\lambda_{k,m}^{(1)} + \lambda_{k,m}^{(2)} + \lambda_{k,m}^{(3)}} \tag{37}$$

where $\kappa_{k,m}$ denotes the planarity measure of the local zone $B_{k,m}$, $\lambda_{k,m}^{(1)}$ is the smallest eigenvalue, reflecting the dispersion of the point cloud along the normal direction, and $\lambda_{k,m}^{(1)} + \lambda_{k,m}^{(2)} + \lambda_{k,m}^{(3)}$ represents the total spatial variance of the local point cloud. When the local point set is closer to a planar structure, the variance along the normal direction becomes smaller, and thus $\kappa_{k,m}$ decreases. Conversely, when the point set exhibits a scattered, volumetric, or multi-structure mixed distribution, the proportion of the smallest eigenvalue increases, leading to a larger $\kappa_{k,m}$.

Therefore, a planarity threshold $\tau_\kappa$ is set, and the following constraint is adopted: $\kappa_{k,m} \le \tau_\kappa$, where $\tau_\kappa$ denotes the maximum allowable local curvature threshold. Only local zones satisfying this condition are regarded as reliable local planar structures and are passed to the subsequent normal angle constraint and point-to-plane distance criterion. This step can effectively filter out non-planar regions composed of vegetation, obstacle edges, or sparse outliers, thereby reducing the risk of non-ground point misclassification caused by erroneous plane fitting.

### 3.2.5 Normal Angle Constraint

Although the planarity constraint can filter out local zones with irregular spatial distributions, some non-ground structures, such as walls, vehicle side surfaces, or local surfaces of pole-like objects, may also exhibit strong planarity. To further distinguish ground planes from vertical or inclined non-ground planes, a normal angle constraint is introduced in this paper.

Let the unit vector in the vertical direction be denoted as:

$$e_z = \left[0,0,1\right]^T \tag{38}$$

Since the direction of the plane normal has sign ambiguity, i.e., $n_{k,m}$ and $-n_{k,m}$ represent the same plane, the angle between the local plane normal and the vertical direction is computed using the absolute inner product:

$$\phi_{k,m} = \arccos\left(\left|n_{k,m}^T e_z\right|\right) \tag{39}$$

where $\phi_{k,m}$ denotes the included angle between the local plane normal and the vertical direction. When the local plane is close to the horizontal ground surface, its normal vector is close to the vertical direction, and $\phi_{k,m}$ is small. When the local plane corresponds to a wall, vehicle side surface, or other vertical structure, its normal vector is closer to the horizontal direction, and $\phi_{k,m}$ becomes larger.

A normal angle threshold $\tau_\phi$ is therefore set, and the following constraint is adopted:

$$\phi_{k,m} \le \tau_\phi \tag{40}$$

where $\tau_\phi$ is used to control the allowable slope range of acceptable ground surfaces. This constraint can effectively suppress the misclassification of vertical structures or steep non-ground planes as ground while allowing road slopes and local undulations.

### 3.2.6 Ground Candidate Extraction Based on the Point-to-Plane Distance Criterion

When the local zone simultaneously satisfies the planarity constraint and the normal angle constraint, the estimated plane can be considered to have high ground candidate confidence. Subsequently, the orthogonal distance from all points in this zone to the local plane $\Pi_{k,m}$ is computed as follows:

$$d_i = \left|n_{k,m}^T\left(\mathrm{X}_i - \mu_{k,m}\right)\right|, p_i \in B_{k,m} \tag{41}$$

where $d_i$ denotes the absolute orthogonal distance from point $p_i$ to the local plane $\Pi_{k,m}$. If a point is close to the local ground plane, it is more likely to belong to the ground region. Conversely, if the distance is relatively large, the point may correspond to vehicles, pedestrians, curbs, vegetation, or other non-ground structures.

Let the point-to-plane distance threshold be $\tau_d$. The candidate ground point set in the k -th ring and the m -th sector is then defined as:

$$G_{k,m}^c = \left\{p_i \in B_{k,m} \left| d_i \le \tau_d \right.\right\} \tag{42}$$

where $G_{k,m}^c$ denotes the candidate ground point set extracted from the local zone $B_{k,m}$, and $\tau_d$ represents the maximum allowable point-to-plane distance. Finally, the candidate ground point sets from all valid local zones are merged to obtain the coarse segmentation result:

$$G^c = \bigcup_{k=1}^{K}\bigcup_{m=1}^{M_k} G_{k,m}^c \tag{43}$$

where $G^c$ denotes the global candidate ground point set obtained in the coarse segmentation stage. It should be noted that $G^c$ is not the final ground point set, but serves as the input to the subsequent fine segmentation stage. The coarse segmentation stage mainly extracts points that may belong to the ground based on local geometric consistency, and therefore can better preserve ground point recall. However, in complex mixed regions, a small number of non-ground points may still be incorrectly classified as candidate ground points. To further improve the reliability of the segmentation results, the subsequent fine segmentation stage introduces reflectance intensity consistency and local height stability to perform secondary discrimination on the candidate points.

## 3.3 Fine Segmentation Based on Intensity Consistency and Local Height Stability for Ground Point Refinement

After the coarse segmentation stage, the candidate ground point set $G^c$ is obtained. This set is mainly extracted based on geometric constraints such as local planarity, normal angle, and point-to-plane distance, and thus can effectively preserve ground point recall. However, in complex road scenarios, some non-ground structures may be close to the local ground plane, such as vehicle underbodies, low curbs, speed bumps, sparse vegetation, and small obstacles. These points may therefore be incorrectly classified as candidate ground points. Meanwhile, due to variations in ground material, point cloud sparsity, and local plane fitting errors, some true ground points may also exhibit weak geometric consistency. To further improve the reliability of the segmentation results, a fine segmentation stage is designed based on the coarse segmentation results. Specifically, a reflectance intensity consistency constraint is used to distinguish high-confidence ground points from uncertain points, and the local height stability of high-confidence neighborhoods is further used to perform secondary discrimination on uncertain points. Finally, robust height statistics are adopted to remove misclassified high-elevation points.

### 3.3.1 2D Neighborhood Construction

Since ground segmentation mainly focuses on the spatial continuity of points in local horizontal regions, a horizontal 2D neighborhood is used for local consistency analysis in the fine segmentation stage. For any candidate ground point $p_i \in G^c$, its horizontal coordinate vector is defined as:

$$q_i = [x_i, y_i]^T \in \mathbb{R}^2 \tag{44}$$

where $q_i$ denotes the 2D coordinates of point $p_i$ on the horizontal xOy -plane, and $x_i$ and $y_i$ are the horizontal coordinate components of this point in the LiDAR coordinate system, respectively. Compared with a 3D Euclidean neighborhood, the 2D horizontal neighborhood can more directly characterize the local spatial continuity of the ground region, preventing non-ground points with large height differences from being excessively involved in ground neighborhood statistics due to their close 3D distances.

Given the intensity consistency analysis radius $\rho_I$, the local candidate neighborhood of the candidate point $p_i$ is defined as:

$$\mathcal{N}_i^I = \left\{ p_j \in G^c \,\middle|\, \left\| q_j - q_i \right\|_2 \le \rho_I, j \ne i \right\} \tag{45}$$

where $\mathcal{N}_i^I$ denotes the 2D intensity neighborhood of point $p_i$ within the candidate ground point set $G^c$, $\rho_I$ is the neighborhood radius for the intensity consistency constraint, $\|\cdot\|_2$ represents the Euclidean norm, $p_j$ denotes a candidate point within the neighborhood, and $j \ne i$ indicates that the point itself is excluded from the neighborhood statistics. This neighborhood is used to evaluate the reflectance intensity consistency between the candidate point and its surrounding candidate ground points.

### 3.3.2 Reflectance Intensity Consistency Constraint

Within a local ground region, the material and surface structure usually exhibit certain continuity. Therefore, the reflectance intensity of neighboring ground points changes relatively smoothly. In contrast, vehicles, curbs, guardrails, vegetation, and other obstacles usually have different reflection characteristics, which can introduce large intensity fluctuations within local neighborhoods. Thus, reflectance intensity consistency can serve as an auxiliary criterion beyond geometric constraints to further distinguish reliable ground points from potential misclassified points.

For a candidate point $p_i \in G^c$, its average intensity difference within the neighborhood is defined as:

$$\Delta I_i = \frac{1}{\left|\mathcal{N}_i^I\right|} \sum_{p_j \in \mathcal{N}_i^I} \left| I_i - I_j \right|, \left|\mathcal{N}_i^I\right| \ge N_I^{\min} \tag{46}$$

where $\Delta I_i$ denotes the average reflectance intensity difference between point $p_i$ and its neighboring candidate points. $I_i$ and $I_j$ represent the reflectance intensities of point $p_i$ and neighboring point $p_j$, respectively, and $\left| I_i - I_j \right|$ denotes the absolute intensity difference between the two points. $\mathcal{N}_i^I$ denotes the number of neighboring points, and $N_I^{\min}$ represents the minimum number of neighboring points required for intensity consistency computation. The purpose of introducing $N_I^{\min}$ is to avoid unreliable intensity statistics when the number of neighboring points is insufficient.

Let the intensity consistency threshold be $\tau_I$. Then, the high-confidence ground point set is defined as:

$$G^h = \left\{ p_i \in G^c, \left\|\mathcal{N}_i^I\right| \ge N_I^{\min} \Delta I_i \le \tau_I \right\} \tag{47}$$

where $G^h$ denotes the high-confidence ground point set, and $\tau_I$ represents the maximum allowable average intensity difference. When $\Delta I_i \le \tau_I$, point $p_i$ exhibits good reflectance intensity consistency with the local candidate ground neighborhood, and thus can be regarded as a high-confidence ground point.

Correspondingly, the uncertain point set is defined as:

$$G^u = G^c \setminus G^h \tag{48}$$

where $G^u$ denotes the set of candidate points that fail to satisfy the intensity consistency constraint or have insufficient neighboring points. It should be noted that uncertain points are not directly classified as non-ground points. This is because true ground regions may also exhibit intensity discontinuities due to abrupt material changes, water accumulation, surface wear, shadows, incidence angle variations, or sensor echo fluctuations. If all uncertain points are directly removed, ground under-segmentation may easily occur. Therefore, a local height stability constraint is further introduced to perform secondary discrimination on uncertain points.

#### 3.3.3 Ground Point Refinement Based on Local Height Stability in High-Confidence Neighborhoods

For an uncertain point $p_i \in G^u$, if it truly belongs to the ground, its height should remain locally continuous with the surrounding high-confidence ground points. Conversely, if it corresponds to a vehicle underbody, a point above a curb, or a low obstacle, its height will usually disrupt the local height stability of the ground surface. Therefore, high-confidence ground points are used as reliable references in this paper. A high-confidence neighborhood is constructed for each uncertain point, and local height stability is used to determine whether the uncertain point should be restored as a ground point.

Given the height refinement neighborhood radius $\rho_H$, the high-confidence neighborhood of an uncertain point $p_i \in G^u$ is defined as:

$$\mathcal{N}_i^H = \left\{ p_j \in G^h \,\middle|\, \left\| q_j - q_i \right\|_2 \le \rho_H \right\} \tag{49}$$

where $\mathcal{N}_i^H$ denotes the high-confidence ground neighborhood around the uncertain point $p_i$, $\rho_H$ represents the neighborhood radius for height stability refinement, and $q_i$ and $q_j$ denote the horizontal coordinates of points $p_i$ and $p_j$, respectively. This neighborhood consists only of high-confidence ground points, and therefore provides more reliable local height references than directly using all candidate points.

When $|\mathcal{N}_i^H| \ge N_H^{\min}$, the mean height of the high-confidence neighborhood is computed as:

$$\bar{z}_i^H = \frac{1}{\left|\mathcal{N}_i^H\right|} \sum_{p_j \in \mathcal{N}_i^H} z_j \tag{50}$$

where $\bar{z}_i^H$ denotes the mean height of the high-confidence ground neighborhood around the uncertain point $p_i$, $z_j$ represents the vertical coordinate of the high-confidence neighboring point $p_j$, and $N_H^{\min}$ denotes the minimum number of high-confidence neighboring points required for height stability evaluation.

To determine whether the uncertain point is height-continuous with the local ground surface, the uncertain point $p_i$ and its high-confidence neighborhood are jointly used to construct a local height sample set:

$$\mathcal{Z}_i^H = \{z_i\} \cup \left\{ z_j \,\middle|\, p_j \in \mathcal{N}_i^H \right\} \tag{51}$$

where $\mathcal{Z}_i^H$ denotes the local height sample set composed of the height of the uncertain point and the heights of points in its high-confidence neighborhood, and $z_i$ represents the height of the uncertain point $p_i$. The reason for including $z_i$ in the height sample set is that the joint height distribution remains stable only when the uncertain point is height-consistent with the neighboring high-confidence ground points. If the point is non-ground, its height deviation will increase the local height dispersion.

The local height stability indicator is further defined as:

$$\sigma_i^H = \sqrt{\frac{1}{\left|\mathcal{Z}_i^H\right|} \sum_{z \in \mathcal{Z}_i^H} \left( z - \bar{z}_i^* \right)^2} \tag{52}$$

where $\sigma_i^H$ denotes the local height standard deviation jointly formed by the uncertain point $p_i$ and its high-confidence neighborhood, $\left|\mathcal{Z}_i^H\right|$ represents the number of local height samples, and $\bar{z}_i^*$ denotes the mean height of the set $\mathcal{Z}_i^H$, which is defined as:

$$\bar{z}_i^* = \frac{1}{\left|\mathcal{Z}_i^H\right|} \sum_{z \in \mathcal{Z}_i^H} z \tag{53}$$

Eq. (53) is used to measure the dispersion of the local height distribution after an uncertain point is incorporated into its high-confidence ground neighborhood. If $\sigma_i^H$ is small, the uncertain point is height-continuous with the surrounding high-confidence ground points and is therefore more likely to belong to the true ground. Conversely, if $\sigma_i^H$ is large, the point disrupts the local height stability of the ground surface and is more likely to correspond to a non-ground structure.

Let the local height stability threshold be $\tau_H$. Then, the set of uncertain points that can be restored as ground points is defined as:

$$G^{r,u} = \left\{ p_i \in G^u \,\middle\|\, \left|\mathcal{N}_i^H\right| \geq N_H^{\min}, \sigma_i^H \leq \tau_H \right\} \tag{54}$$

where $G^{r,u}$ denotes the ground point set restored from the uncertain point set, and $\tau_H$ represents the maximum allowable local height standard deviation. When an uncertain point satisfies Eq. (54), it indicates that the point remains height-stable with its high-confidence ground neighborhood. Therefore, it is re-incorporated into the ground point set.

Accordingly, the intermediate ground point set obtained after intensity consistency-based classification and height stability refinement is defined as:

$$G^r = G^h \cup G^{r,u} \tag{55}$$

where $G^r$ denotes the refined ground point set that has not yet undergone one-step robust height removal. This refinement strategy only allows uncertain points to be restored as ground points when they satisfy the height stability condition, and does not continue to expand new ground points from non-candidate points. In this way, it can alleviate ground under-segmentation caused by intensity fluctuations while limiting the spread of false detections, thereby improving the stability of the fine segmentation stage.

#### 3.3.4 Secondary Removal of Misclassified High-Elevation Points Based on Robust Height Statistics

Although most ground points can be correctly preserved after intensity consistency-based classification and local height stability refinement, a small number of high-elevation misclassified points may still remain near vehicle underbodies, curbs, or low obstacles. If the mean and variance are directly used for global height statistics, these outliers may bias the estimated ground height mean and variance, weakening the constraining effect of height statistics. Therefore, a robust height statistics strategy based on the median and median absolute deviation is further adopted in this paper to secondarily remove misclassified high-elevation points.

First, the height sample set is extracted from the intermediate ground point set $G^r$:

$$\mathcal{Z}^r = \left\{ z_i \,\middle|\, p_i \in G^r \right\} \tag{56}$$

where $\mathcal{Z}^r$ denotes the height set corresponding to the refined ground point set $G^r$, and $z_i$ represents the vertical coordinate of point $p_i$.

The height median is computed as:

$$z_{\text{med}} = \operatorname{median}\left(\mathcal{Z}^r\right) \tag{57}$$

where $z_{\text{med}}$ denotes the height median of the set $\mathcal{Z}^r$, and $\operatorname{median}(\cdot)$ represents the median operation. Compared with the mean, the median is less sensitive to a small number of abnormal high-elevation points and can more stably describe the height center of the main ground points.

The median absolute deviation is further computed as:

$$\text{MAD} = \operatorname{median}\left(\left\{ \left|z_i - z_{\text{med}}\right| \,\middle|\, z_i \in \mathcal{Z}^r \right\}\right) \tag{58}$$

where $\text{MAD}$ denotes the median absolute deviation of the height samples and is used to measure the robust dispersion of the ground height distribution. Compared with the standard deviation, $\text{MAD}$ is less affected by a small number of misclassified high-elevation points, making it more suitable for detecting height outliers in complex road scenarios.

Let the robust height scale factor be $\beta_z$, the $\text{MAD}$ consistency coefficient be $c_{\text{MAD}}$, and the minimum height band be $h_z^{\min}$. Then, the adaptive upper height bound is defined as:

$$T_z = z_{\text{med}} + \max\left(\beta_z c_{\text{MAD}} \text{MAD}, h_z^{\min}\right) \tag{59}$$

where $T_z$ denotes the adaptive upper height bound used to remove misclassified high-elevation points, $\beta_z$ is the height threshold scale factor used to adjust the removal strength, $c_{\text{MAD}}$ is the conversion coefficient from MAD to the standard-deviation scale and is usually set to 1.4826, $h_z^{\min}$ denotes the minimum height band used to prevent an excessively small threshold from incorrectly removing true ground points when ground height variation is limited, and $\max(\cdot)$ represents the maximum operation.

Finally, the final ground point set is defined as:

$$G=\left\{ p_i \in G^r \left| z_i \le T_z \right. \right\} \tag{60}$$

where G denotes the final ground point set. Correspondingly, the non-ground point set can be represented as:

$$N = P \setminus G \tag{61}$$

where $N$ denotes the final non-ground point set, and $P$ is the original point cloud set. Eq. (61) only removes high-elevation points from the refined ground point set and does not further expand the ground region. Therefore, it can suppress misclassified high-elevation points while preserving the completeness of ground points. In summary, the fine segmentation stage adopts a cascaded strategy consisting of intensity consistency filtering, height stability refinement based on high-confidence neighborhoods, and secondary robust height removal, thereby further improving the accuracy and reliability of ground segmentation based on the coarse segmentation candidates.

# 4 Experiments

The proposed method was comprehensively evaluated. Section 4.1 introduces the experimental settings. In Section 4.2, quantitative and visual comparisons are conducted between the proposed method and existing ground segmentation methods on the public SemanticKITTI dataset and self-collected point cloud acquired using a Ruby Plus LiDAR, respectively, to verify its segmentation performance in both public benchmark scenarios and real-world acquisition scenarios. Section 4.3 presents ablation experiments to analyze the contributions of the Adaptive Concentric Zone, the geometric coarse segmentation module, and the intensity-guided fine segmentation module to the overall performance. Section 4.4 analyzes the computational complexity of the proposed method and evaluates its computational efficiency and application potential in practical point cloud processing tasks.

## 4.1 Experimental setup

### 4.1.1 Dataset

In the experimental evaluation, the SemanticKITTI dataset [34] and self-collected point cloud were used to jointly validate the ground segmentation performance of the proposed method. The SemanticKITTI dataset was used for quantitative comparison with existing ground segmentation algorithms. Based on its point-wise semantic labels, the categories of road, parking, sidewalk, other-ground, lane-marking, and terrain were labeled as ground points, while the remaining categories were labeled as non-ground points, thereby constructing binary ground-truth labels for ground segmentation.

To further verify the applicability of the proposed method in real sensor acquisition scenarios, road-scene point cloud were collected using a RoboSense Ruby Plus 128-channel LiDAR. Ruby Plus is a 128-channel LiDAR designed for commercial L4 autonomous driving operation scenarios. It provides strong long-range detection capability, with a detection range of up to 240 m under 10% reflectivity. Meanwhile, the sensor provides a high angular resolution of $0.1^\circ \times 0.1^\circ$ at an operating frequency of 10 Hz, which helps acquire richer information from far-range road environments. The LiDAR also provides suppression capabilities against multi-LiDAR interference and ambient light interference, and is designed for stability and reliability in vehicle-mounted operating environments. Therefore, Ruby Plus is suitable for autonomous vehicles, roadside perception systems, mobile robots, and high-precision 3D environmental perception, as shown in Fig. 3.

In summary, by combining a public dataset with self-collected point cloud, this paper comprehensively verifies the effectiveness and robustness of the proposed ground segmentation method from two aspects: standardized ground-truth evaluation and adaptability to real road scenarios.

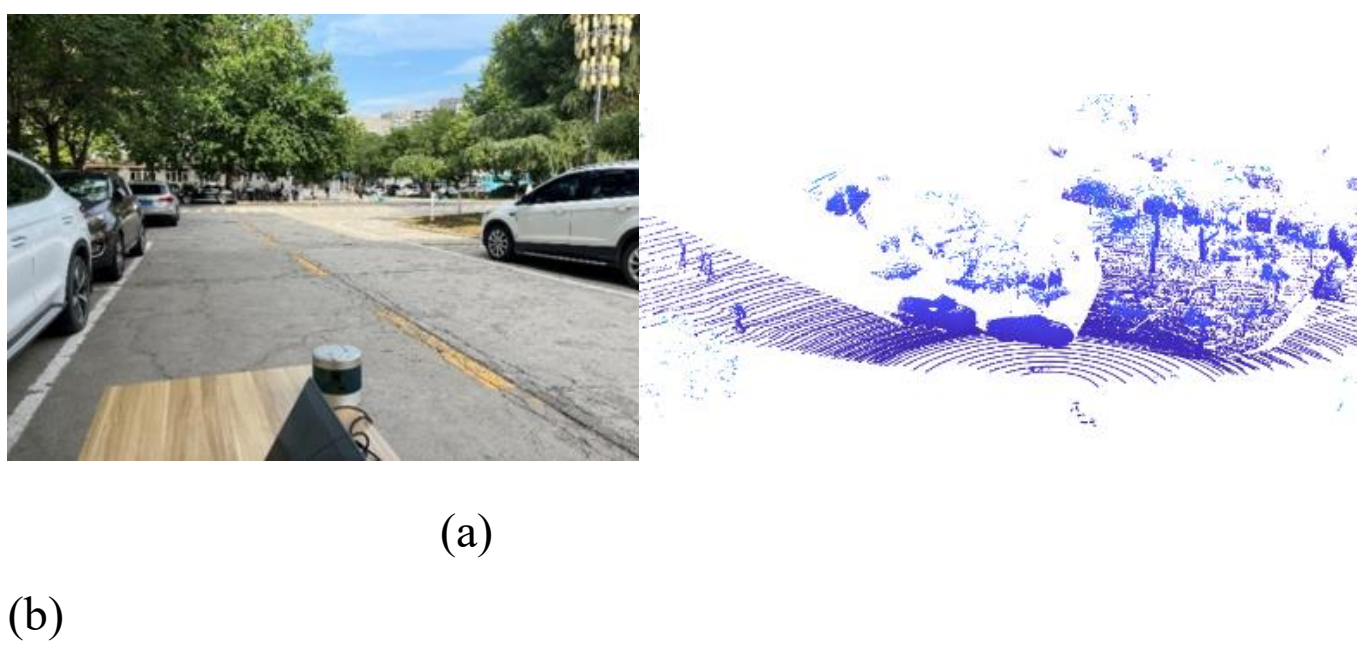

(a)

(b)

Fig. 3 Ruby Plus LiDAR acquisition scenario and its point cloud

Cohen's Kappa, denoted as $\kappa$, is a statistical measure used to evaluate the consistency of annotation results and is commonly adopted for assessing labeling agreement in classification tasks. Unlike the direct calculation of accuracy, the Kappa coefficient corrects for apparent agreement caused by random chance, and therefore provides a more reliable measure of true agreement.

To ensure the reliability of the RUBY PLUS ground point annotations, a strict annotation protocol was formulated and implemented. Three independent LiDAR data experts performed blind annotation on the same point cloud subset. The initial annotation was completed by a senior expert, and was then jointly reviewed and corrected by the other two experts until all uncertain cases were resolved. The final ground truth was determined based on the consensus among the three experts. The statistical results are as follows: the scene contains 230,400 points in total, including 31,857 ground points and 198,543 non-ground points.

If two annotators consistently classify a point as a "ground-point signal", the number of such points is denoted as $N_{agree}^{signal}$. If both annotators classify a point as "non-ground-point noise", the number of such points is denoted as $N_{agree}^{noise}$. For cases in which the two annotators give opposite judgments, the number is denoted as $N_{disagree}$. Accordingly, the observed agreement ratio is denoted as $p_0^c$, and its calculation formula is given as follows:

$$p_0^c = \frac{N_{agree}^{signal} + N_{agree}^{noise}}{N_{agree}^{signal} + N_{agree}^{noise} + N_{disagree}} \tag{62}$$

For annotator A, the proportion of points classified as "ground-point signal" is denoted as $p_A^{signal} = N_A^{signal} / N_{total}$, and the corresponding proportion for annotator B is denoted as $p_B^{signal} = N_B^{signal} / N_{total}$. Similarly, the proportions of points classified by the two annotators as "non-ground-point noise" can be calculated as $p_A^{noise}$ and $p_B^{noise}$, respectively. These probabilities are used to compute the expected agreement $p_e^c$, as shown in Eq. (63).

$$p_e^c = p_A^{signal} \cdot_B^{signal} + p_A^{noise} \cdot_B^{noise} \tag{63}$$

The calculation result of the Kappa coefficient $\kappa$ is shown in Eq. (64):

$$\kappa^c = \frac{p_0^c - p_e^c}{1 - p_e^c} \tag{64}$$

After summarizing the annotation results of annotators A, B, and C, and performing the calculation according to Eqs. (62)–(63), the obtained results are presented in Table 1.

Table 1 Statistical summary of manual annotation results for RUBY PLUS LiDAR point cloud.

| Data | Pairwise | Consistent number of ground points | Number of consistent non-ground points | Number of points of disagreement | $p_0^c$ | $p_e^c$ | $\kappa^c$ |
|---|---|---|---|---|---|---|---|
| Point cloud | A vs B | 30742 | 193413 | 6245 | 0.9729 | 0.7492 | 0.8919 |
| | B vs C | 31318 | 195525 | 3557 | 0.9846 | 0.7539 | 0.9373 |
| | A vs C | 29576 | 192544 | 8280 | 0.9641 | 0.7501 | 0.8562 |

As shown in Table 1, the $\kappa$ values obtained from the consistency tests of the three annotator groups are 0.8919, 0.9373, and 0.8562, respectively, with an average value of 0.8951. According to the evaluation criterion of the Kappa coefficient, a $\kappa$ value within the range of 0.81–1.00 indicates almost perfect agreement. Therefore, the manually annotated "ground point" ground-truth labels constructed for the RUBY PLUS LiDAR point cloud exhibit high consistency and reliability, and can be used for the quantitative evaluation of subsequent ground segmentation algorithms.

#### 4.1.2 Dataset

ollowing previous work [30], points labeled as vegetation in the SemanticKITTI dataset are not evaluated as either ground or non-ground points, because treating vegetation as a single ground or non-ground class is inappropriate. This means that vegetation-labeled points are excluded only during the evaluation stage, while they are still retained in the input point cloud. In this paper, Precision, Recall, and F1-score are adopted as the evaluation metrics, as described in [30].

$$P_d = \frac{N_q}{N_s} \tag{65}$$

$$R_d = \frac{N_q}{N_y} \tag{66}$$

$$F1 = \frac{2P_d R_d}{P_d + R_d} \tag{67}$$

where $N_q$ denotes the number of points correctly segmented as ground points, i.e., the number of points predicted as ground whose ground-truth labels are also ground; $N_s$ denotes the total number of points predicted as ground by the algorithm; and $N_y$ denotes the total number of ground points in the ground-truth labels. Therefore, $P_d$ represents the precision of ground segmentation, namely the proportion of true ground points among the predicted ground points; $R_d$ represents the recall of ground segmentation, namely the proportion of true ground points that are correctly detected; and F1-score is the harmonic mean of precision and recall, which is used to comprehensively evaluate the ground segmentation performance.

#### 4.1.3 Parameter Settings

Table 2 presents the key parameter settings of the proposed two-stage ground segmentation method based on the Adaptive Concentric Zone Model. The overall parameters are configured according to three modules: Adaptive Concentric Zone, coarse segmentation, and fine segmentation. The Adaptive Concentric Zone module is mainly controlled by the desired number of rings $K_0$, the minimum expected number of points per ring $N_r^{\min}$, the minimum number of rings $K_{\min}$, the target arc length L, the minimum number of sectors $M_{\min}$, and the maximum number of sectors $M_{\max}$. Specifically, $K_0 = 30$ is used to limit the maximum desired number of rings for radial quantile-based ring partitioning. $N_r^{\min} = 1000$ and $K_{\min} = 5$ are used to adaptively determine the effective number of rings K according to the scale of the current point cloud frame, thereby reducing invalid far-range rings while preserving the basic radial partitioning capability. The target arc length $L = 1.2$ m is used to dynamically determine the number of sectors $M_k$ in each ring based on the principle of approximately constant arc length. Meanwhile, $M_{\min} = 12$ and $M_{\max} = 90$ are used to constrain the range of sector numbers, preventing insufficient angular resolution in near-range regions or insufficient points in local zones caused by excessive partitioning in far-range regions.

In the coarse segmentation stage, the minimum number of points per local zone is set to $N_b^{\min} = 10$, ensuring that each local zone $B_{k,m}$ contains sufficient points for reliable local plane estimation and avoiding ill-conditioned fitting caused by inadequate samples. The lowest-height seed band is set to $h_s = 0.40$ m to restrict the range of low-height seed points involved in weighted PCA-based plane fitting. The boundary weight $\eta = 0.05$ specifies the weight decay ratio at a relative height of $h_s$. According to $\alpha = -\ln(\eta) / h_s$, the height-decay coefficient is calculated as $\alpha = 7.49, \mathrm{m}^{-1}$, which controls the contribution of points with relatively large heights to local plane fitting. Furthermore, the planarity threshold $\tau * \kappa = 0.04$, normal angle threshold $\tau_\phi = 20^\circ$, and point-to-plane distance threshold $\tau_d = 0.15$ m jointly constitute the local geometric consistency constraints. These constraints are used to reject non-planar zones, suppress non-ground structures with large inclination angles, and extract the candidate ground point set $G^c$ from valid local zones.

In the fine segmentation stage, the intensity consistency neighborhood radius is set to $\rho_I = 0.50$ m to construct the 2D intensity neighborhood $N_i^I$ for each candidate ground point. The minimum number of neighboring points required for intensity consistency analysis is set to $N_I^{\min} = 5$ to ensure the reliability of local intensity statistics. Candidate points with insufficient neighbors are assigned to the uncertain point set. The intensity consistency threshold $\tau_I = 30$ is used to further divide the candidate ground points into the high-confidence ground point set $G^h$ and the uncertain point set $G^u$. Subsequently, the height stability neighborhood radius $\rho_H = 0.50$ m is used to construct the high-confidence ground neighborhood $N_i^H$ for each uncertain point. The minimum number of neighboring points required for height refinement is set to $N_H^{\min} = 5$, ensuring sufficient neighborhood support for height stability evaluation. The local height stability threshold $\tau_H = 0.12$ m is used to determine whether the height distribution within the high-confidence neighborhood is stable and whether an uncertain point should be restored as a ground point. These parameter settings reduce the risk of misclassifying non-ground points in locally mixed structures while preserving ground-point recall.

Table 2 Statistical summary of manual annotation results for RUBY PLUS LiDAR point cloud.

| Module | Parameter | Symbol | Value | Note |
|---|---|---|---|---|
| Adaptive Concentric Zone Model | Desired number of rings | $K_0$ | 30 | controls the maximum desired number of rings in radial quantile-based ring partitioning |
| | Minimum expected number of points per ring | $N_r^{\min}$ | 1000 | used to adaptively determine the effective number of rings $K$ according to the number of points in the current frame, thereby reducing invalid far-range rings |
| | Minimum number of rings | $K_{\min}$ | 5 | ensures the basic radial partitioning capability of the point cloud |
| | Target arc length | $L$ | 1.2 m | Used to determine the number of sectors $M_k$ in each ring according to the principle of |

| | | | | approximately constant arc length |
|---|---|---|---|---|
| | Minimum number of sectors | $M_{\min}$ | 12 | Ensures that each ring maintains a basic angular resolution |
| | Maximum number of sectors | $M_{\max}$ | 90 | Prevents excessive subdivision of far-range rings, which may result in insufficient points within local zones. |
| | Minimum number of points per local zone | $N_b^{\min}$ | 10 | Ensures that each local zone $B_{k,m}$ contains sufficient points for local plane estimation |
| | Lowest-height seed band width | $h_s$ | 0.40 m | Restricts the range of low-height seed points involved in weighted PCA-based plane fitting. |
| Coarse Segmentation | Seed-band boundary weight | $\eta$ | 0.05 | Specifies the weight decay ratio when the point height reaches $h_s$ |
| | Height-decay coefficient | $\alpha$ | $7.49m^{-1}$ | Calculated as $\alpha = -\ln(\eta) / h_s$ and used to control the height-decay weighting |
| | Planarity threshold | $\tau_\kappa$ | 0.04 | Used to filter out non-planar local zones with relatively large curvature |
| | Normal angle threshold | $\tau_\phi$ | (20^\circ) | Constrains the angle between the local plane normal and the vertical direction |
| | Point-to-plane distance threshold | $\tau_d$ | 0.15 m | Used to extract the candidate ground point set $G^c$ from valid local zones |
| | Intensity consistency neighborhood radius | $\rho_I$ | 0.50 m | Used to construct the 2D intensity neighborhood $N_i^I$ of each candidate ground point |
| | Minimum number of neighboring points for intensity consistency | $N_I^{\min}$ | 5 | When the number of neighboring points is insufficient, the candidate point is classified as an uncertain point |
| | Intensity consistency threshold | $\tau_I$ | 30 | Used to distinguish the high-confidence ground point set $G^h$ from the uncertain point set $G^u$ |
| Fine Segmentation | Height stability neighborhood radius | $\rho_H$ | 0.50 m | Used to construct the high-confidence ground neighborhood $N_i^H$ for each uncertain point |
| | Minimum number of neighboring points for height refinement | $N_H^{\min}$ | 5 | Ensures sufficient high-confidence neighborhood support for uncertain-point refinement. |
| | Local height stability threshold | $\tau_H$ | 0.12 m | Used to assess the height stability of the high-confidence neighborhood and determine whether an uncertain point should be restored as a ground point |

During parameter tuning, the partition scale was first determined, followed by adjustment of the geometric constraints in the coarse segmentation stage and optimization of the confidence criteria in the fine segmentation stage. For the Adaptive Concentric Zone module, the desired number of rings $K_0$ and the target arc length L jointly determine the scale of the local zones. Increasing $K_0$ or decreasing L produces finer radial and angular partitions, which facilitates the representation of ground structures near road boundaries, local undulations, and far-range sparse regions. However, excessively fine partitioning may result in insufficient points within individual local zones, leading to unstable PCA-based plane fitting and increased computational overhead. Conversely, decreasing $K_0$ or increasing L improves the point-count stability of local plane fitting but may group ground points together with non-ground structures, such as vehicles, vegetation, and curbs, within the same zone. This can bias local plane estimation and consequently increase the misclassification of non-ground points. Therefore, $K_0$, L, $M_{\min}$, and $M_{\max}$ should be balanced between local partition resolution and point-count stability. In the implementation, $K_0 = 30$, $N_r^{\min} = 1000$, $L = 1.2,\mathrm{m}$, $M_{\min} = 12$, and $M_{\max} = 90$ are adopted to maintain relatively balanced local modeling scales across regions at different ranges.

In the coarse segmentation stage, the point-to-plane distance threshold $\tau_d$, normal angle threshold $\tau_\phi$, and planarity threshold $\tau_\kappa$ are the key parameters affecting Precision and Recall. Increasing $\tau_d$ relaxes the ground candidate criterion, allowing more points to be classified as ground and generally improving Recall. However, it also increases the risk of misclassifying non-ground points, such as vehicle underbodies, curbs, and low obstacles, as ground, thereby reducing Precision. Conversely, decreasing $\tau_d$ suppresses non-ground point misclassification but may cause low-lying ground points, sloped ground surfaces, or sparse far-range ground points to be missed. Similarly, increasing $\tau_\phi$ or $\tau_\kappa$ relaxes the orientation and shape constraints on local planes, which helps preserve ground points in regions with relatively steep slopes or pronounced local irregularities, but may also introduce walls, vehicle side surfaces, vegetation, and other non-ground structures. Reducing these thresholds improves the reliability of plane estimation and the suppression of non-ground points, but decreases adaptability to complex terrain. The lowest-height seed band $h_s$ and the height-decay coefficient $\alpha$ primarily affect the stability of weighted PCA-based plane fitting. A relatively large $h_s$ includes more candidate points in the fitting process, which improves the stability of local plane estimation but may introduce interference from elevated non-ground points. A relatively small $h_s$ strengthens the dominant contribution of low-height ground seeds, but may result in insufficient fitting samples in sparse regions. In this study, $\tau_d = 0.15,\mathrm{m}$, $\tau_\phi = 20^\circ$, $\tau_\kappa = 0.04$, and $h_s = 0.40,\mathrm{m}$ are adopted. With the boundary weight set to $\eta = 0.05$, the corresponding height-decay coefficient is calculated as $\alpha = 7.49,\mathrm{m}^{-1}$, achieving a balance between ground point completeness and non-ground point suppression.

In the fine segmentation stage, the intensity consistency threshold $\tau_I$, neighborhood radii $\rho_I$ and $\rho_H$, minimum neighborhood sizes $N_I^{\min}$ and $N_H^{\min}$, and height stability threshold $\tau_H$ jointly determine the secondary classification of uncertain points. Increasing $\tau_I$ relaxes the intensity consistency constraint, allowing more candidate points to be classified as high-confidence ground points and potentially improving Recall. However, when local non-ground structures are mixed with ground points, this may also increase the risk of misclassification. Conversely, decreasing $\tau_I$ makes the selection of high-confidence ground points more stringent, which is beneficial for improving Precision, but may classify some true ground points as uncertain points. If the neighborhood radii $\rho_I$ and $\rho_H$ are too small, insufficient neighboring points may lead to unreliable intensity statistics and height stability evaluation. Excessively large neighborhood radii, however, may cross road boundaries or encompass regions with different structures, thereby weakening the discriminative capability of local consistency constraints. Increasing $\tau_H$ makes uncertain points more likely to be restored as ground points, which may improve Recall but also increase the risk of misclassification. Decreasing $\tau_H$ results in a more conservative refinement process, helping to suppress non-ground point misclassification but potentially retaining some missed ground points. In this study, $\tau_I = 30$, $\rho_I = \rho_H = 0.50,\text{m}$, $N_I^{\min} = N_H^{\min} = 5$, and $\tau_H = 0.12,\text{m}$ are adopted. These settings enable the intensity consistency constraint and local height stability refinement to preserve ground point completeness while further suppressing the misclassification of non-ground structures.

Overall, when the experimental results contain many blue false-negative points or exhibit low Recall, $\tau_d$, $\tau_\phi$, or $\tau_H$ should be moderately increased, or the local partition scale should be reduced, to improve the preservation of true ground points. Conversely, when many red false-positive points are observed or Precision is low, $\tau_d$, $\tau_\phi$, $\tau_\kappa$, or $\tau_I$ should be appropriately reduced, while the neighborhood radii should be constrained to strengthen the suppression of non-ground structures. Based on the above parameter-tuning strategy, a parameter configuration with favorable generalization capability was determined, achieving a balanced trade-off between ground point recall and non-ground point misclassification. It should be noted that this parameter configuration does not require frame-specific adjustment. The same parameter settings achieved satisfactory segmentation performance on both the public SemanticKITTI dataset and the self-collected RUBY PLUS LiDAR point cloud, demonstrating their strong stability and applicability.

## 4.2 Experimental Results and Discussion

### 4.2.1 Experiments on the SemanticKITTI Dataset

The overall performance of the proposed method and other ground segmentation methods on the SemanticKITTI dataset is summarized in Fig. 4 and Table 3. Fig. 4 presents visual comparisons of the proposed method with several state-of-the-art ground segmentation methods, including DBSCAN, TRAVEL, GroundGrid, FugSeg, PatchWork, and PatchWork++, on different point cloud frames. Green points denote correctly identified ground points, red points represent non-ground points misclassified as ground, and blue points indicate ground points incorrectly classified as non-ground. The compared methods exhibit noticeable performance differences in complex road scenarios. DBSCAN is highly sensitive to variations in point cloud density and produces numerous blue misclassified points near road boundaries, in sparse regions, and over locally undulating surfaces. This indicates that DBSCAN tends to misclassify true ground points as non-ground, resulting in discontinuous ground structures. TRAVEL, GroundGrid, and PatchWork can extract the principal road regions, but still produce varying numbers of red and blue misclassified points in locally non-flat areas, along road boundaries, and near obstacles, indicating limited adaptability to complex terrain variations and local geometric disturbances. FugSeg and PatchWork++ achieve relatively better overall segmentation performance; however, missed ground points and misclassified non-ground points remain in some frames. In particular, the segmentation boundaries are still unstable in regions with sparse point distributions, pronounced ground-height variations, or complex local structures.

In contrast, the proposed method produces more complete and spatially continuous ground segmentation results across all four test point cloud frames. As shown in the figure, the green regions obtained by the proposed method are more continuous, while fewer red and blue misclassified points are observed. This indicates that the proposed method effectively suppresses the misclassification of non-ground points while maintaining high completeness in ground point detection. This improvement is primarily attributed to the Adaptive Concentric Zone, which enhances adaptability to range-dependent variations in point cloud density and improves the stability of ground structure modeling in complex road scenarios. Consequently, the proposed method exhibits stronger robustness and generalization capability in scenes characterized by non-uniform point cloud density, local terrain undulations, and varying road boundaries.

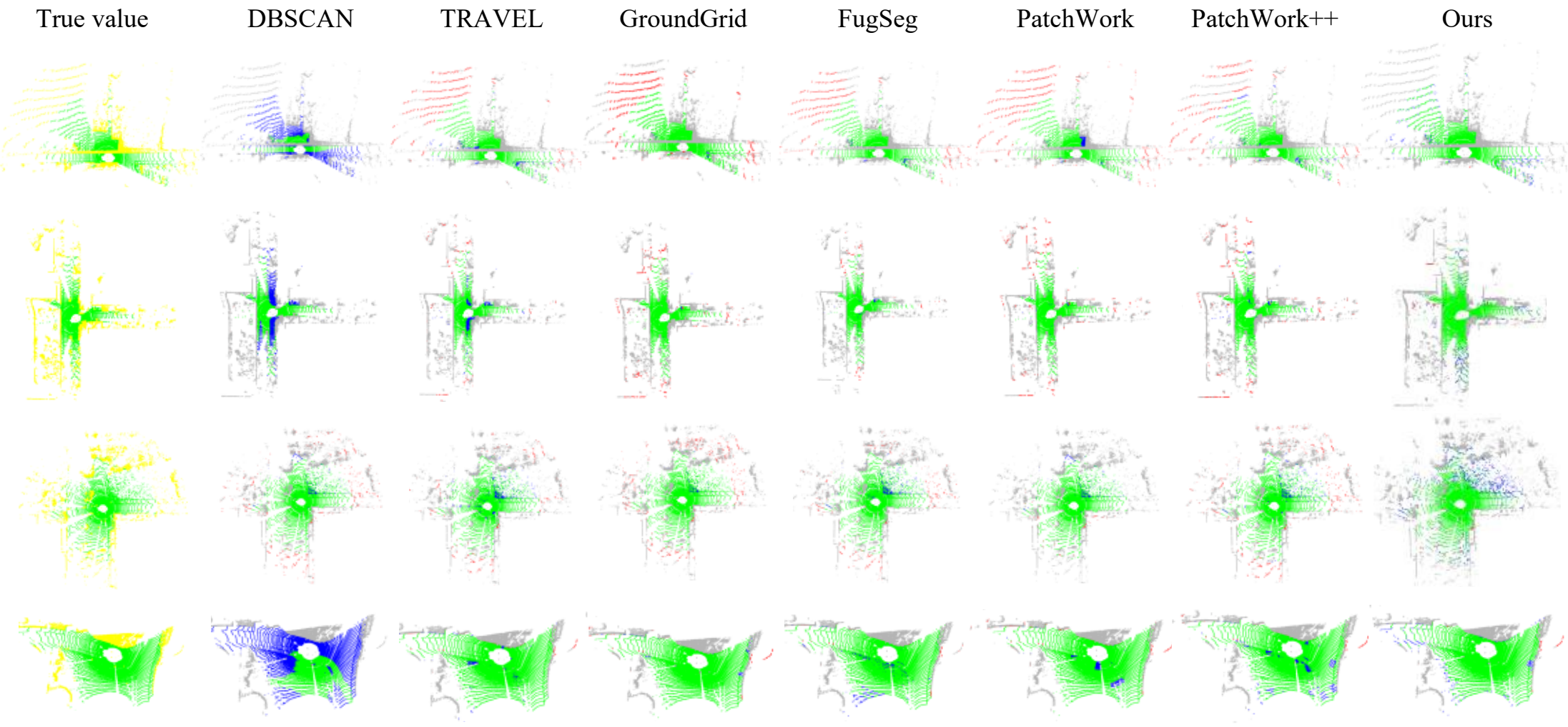

Fig. 4 Qualitative comparison between the proposed method and other methods on (a) frame 504 of sequence 03, (b) frame 501 of sequence 08, (c) frame 100 of sequence 08, and (d) frame 3,300 of sequence 02. The proposed method effectively alleviates under-segmentation in several regions. Green, red, and blue points represent TP, FP, and FN, respectively.

Better performance is indicated by fewer red and blue points and more green points. Best viewed in color

Table 3 further reports the quantitative results of different methods on the test point clouds using Precision, Recall, and F1-score, where Precision and Recall are presented as mean ± standard deviation. As shown in the table, DBSCAN achieves a Precision of 99.14%, which is comparable to that of the proposed method. However, its Recall is only 80.35%, resulting in an F1-score of 88.76%. This indicates that although DBSCAN effectively suppresses the false detection of non-ground points, it misses a large number of true ground points and therefore provides insufficient segmentation completeness. TRAVEL, GroundGrid, and PatchWork achieve F1-scores of 96.18%, 95.24%, and 95.68%, respectively, showing clear improvements over DBSCAN, although limitations remain in either Precision or Recall. FugSeg and PatchWork++ achieve F1-scores of 96.84% and 96.49%, respectively, demonstrating strong overall performance in ground segmentation. Nevertheless, their Precision values are 95.98% and 95.59%, respectively, both lower than that of the proposed method, indicating that a certain proportion of non-ground points are still misclassified as ground.

The proposed method achieves a Precision of 99.12%, a Recall of 96.24%, and an F1-score of 97.66%, yielding the highest overall F1-score among all compared methods. Compared with DBSCAN, the proposed method reduces Precision by only 0.02 percentage points, while improving Recall by 15.89 percentage points and F1-score by 8.90 percentage points. This demonstrates that the proposed method substantially alleviates the missed detection of ground points while maintaining high segmentation precision. Compared with FugSeg and PatchWork++, the proposed method improves F1-score by 0.82 and 1.17 percentage points, respectively, while increasing Precision by 3.14 and 3.53 percentage points. These results indicate that the proposed method more effectively suppresses the misclassification of non-ground points. In addition, the standard deviation of Precision for the proposed method is only 0.29%, lower than that of all other compared methods, indicating smaller performance fluctuations across different sequences and point cloud frames and thus better stability.

Combining the qualitative visualization results with the quantitative evaluation metrics, the proposed method can distinguish ground and non-ground points more accurately in complex road scenarios and achieves a favorable balance among Precision, Recall, and F1-score. The experimental results demonstrate that the proposed method not only maintains high ground point recognition accuracy but also effectively reduces both missed detections and false detections, thereby providing greater robustness and practical value for ground segmentation in complex traffic environments.

Table 3 Performance comparison between the proposed method and other ground segmentation methods on the SemanticKITTI dataset. Metrics are reported as mean ± standard deviation

| Algorithm | Precision (%) | Recall (%) | F1 (%) |
|---|---|---|---|
| DBSCAN | 99.14±0.50 | 80.35±5.58 | 88.76 |
| TRAVEL | 97.02±0.59 | 95.35±3.28 | 96.18 |
| GroundGrid | 95.66±1.73 | 94.82±0.31 | 95.24 |
| FugSeg | 95.98±0.54 | 97.72±1.29 | 96.84 |
| PatchWork | 94.29±0.95 | 97.11±3.67 | 95.68 |
| PatchWork++ | 95.59±3.50 | 97.41±2.83 | 96.49 |
| Ours | 99.12±0.29 | 96.24±1.05 | 97.66 |

#### 4.2.2 Experiments on Self-Collected RUBY PLUS LiDAR Point Cloud

To further evaluate the applicability and robustness of the proposed method in real-world LiDAR acquisition scenarios, ground segmentation experiments were conducted on self-collected point cloud acquired using a RoboSense Ruby Plus 128-channel LiDAR. Compared with public datasets, self-collected point clouds more directly reflect sensor characteristics, point-density distributions, ground undulations, and interference from roadside non-ground structures in actual road environments. In the experiments, ground-truth labels were first generated through manual annotation, and the reliability of the annotations was verified using Cohen’s kappa coefficient. The proposed method was then quantitatively and visually compared with representative ground segmentation methods, including DBSCAN, TRAVEL, GroundGrid, FugSeg, PatchWork, and PatchWork++. Precision, Recall, and F1-score were adopted to comprehensively evaluate the segmentation performance of the different methods, thereby validating the effectiveness of the proposed method in real-world road point cloud scenarios.

Fig. 5 presents the visual ground segmentation results of different methods on the self-collected RUBY PLUS LiDAR point cloud. Green points denote correctly identified ground points, gray points denote correctly identified non-ground points, red points represent non-ground points misclassified as ground, and blue points represent ground points misclassified as non-ground. According to the ground truth, the ground region in this point cloud frame is mainly distributed over the road ahead of the sensor and its extensions on both sides, whereas non-ground structures are primarily located near road boundaries, vegetation, and roadside obstacles. The DBSCAN result contains extensive blue regions, particularly over the main road surface and in far-range sparse areas. This indicates that, under fixed density thresholds, DBSCAN cannot effectively accommodate the range-dependent density distribution of vehicle-mounted LiDAR point clouds, causing many true ground points to be misclassified as non-ground and substantially disrupting ground continuity. TRAVEL produces relatively continuous green regions and successfully recovers most of the road surface. However, a small number of red points remain near roadside non-ground structures, indicating its limited capability to suppress non-ground points in locally mixed regions. GroundGrid preserves a certain degree of continuity over the main road surface, but numerous red points appear near road boundaries and non-ground structures, suggesting that elevation grid-based discrimination may incorrectly classify some non-ground points as ground in regions with complex local structures. Both red and blue points are observed in the FugSeg result, indicating that false positives and false negatives remain while ground recall is preserved, particularly in locally undulating and sparse regions where the classification stability is limited. PatchWork and PatchWork++ preserve the overall ground region more effectively than the preceding methods, with relatively complete green coverage. Nevertheless, blue false-negative points and red false-positive points remain visible near road boundaries and in locally sparse regions, suggesting that fixed or semi-adaptive partitioning may still bias local plane estimation in mixed-structure regions. In contrast, the proposed method produces the most spatially continuous distribution of green points and shows high agreement with the ground-truth ground region. Meanwhile, only a small number of red false-positive points are observed, and blue false-negative points are barely visible. These results indicate that the proposed method effectively suppresses the misclassification of non-ground points while preserving the completeness of true ground points. The results further demonstrate the adaptability of the Adaptive Concentric Zone to non-uniform point cloud density and confirm the effectiveness of the combined coarse segmentation and fine segmentation mechanism in improving the robustness and stability of ground segmentation in complex road scenarios.

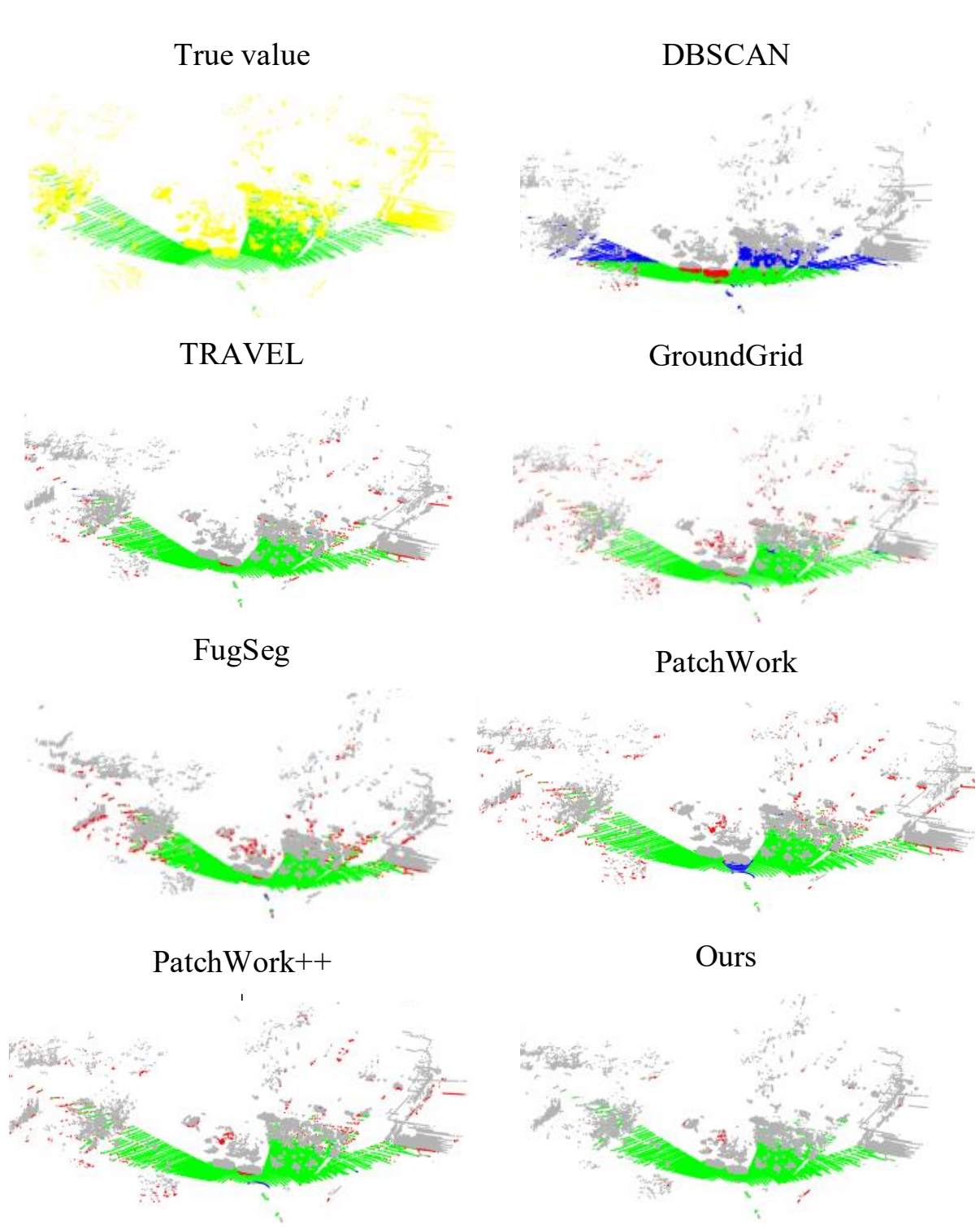

Fig. 5 Visual comparison of ground segmentation results on self-collected RUBY PLUS LiDAR point cloud

Table 4 presents the quantitative evaluation results of different ground segmentation methods on the self-collected RUBY PLUS LiDAR point cloud. The proposed method achieves a Precision of 98.72%, a Recall of 100.00%, and an F1-score of 99.36%, attaining the best or near-best performance across all evaluation metrics. In particular, the Recall of 100.00% indicates that the proposed method fully preserves the true ground points and provides strong ground point retrieval capability. The Precision of 98.72% indicates that only a small proportion of non-ground points are misclassified as ground. Compared with the best-performing baseline methods, the proposed method improves the F1-score by 2.48 percentage points over TRAVEL and by 3.59 percentage points over PatchWork++, demonstrating a more favorable balance between Precision and Recall. Taken together, the visualization results and quantitative metrics show that the proposed two-stage ground segmentation method based on the Adaptive Concentric Zone Model effectively accommodates point cloud sparsity, local ground undulations, and interference from non-ground structures in real-world road environments, thereby achieving high segmentation accuracy and robustness.

Table 4 Performance comparison between the proposed method and other ground segmentation methods on self-collected point cloud acquired using a RUBY PLUS LiDAR

| Algorithm | Precision (%) | Recall (%) | F1 (%) |
|---|---|---|---|
| DBSCAN | 8.97 | 53.82 | 15.38 |
| TRAVEL | 95.47 | 98.33 | 96.88 |
| GroundGrid | 90.78 | 95.09 | 92.89 |
| FugSeg | 87.72 | 99.79 | 93.37 |
| PatchWork | 92.97 | 93.51 | 93.24 |
| PatchWork++ | 95.02 | 96.54 | 95.77 |
| Ours | 98.72 | 100.00 | 99.36 |

The experimental results on both the public SemanticKITTI dataset and the self-collected RUBY PLUS LiDAR point cloud demonstrate that the proposed two-stage ground segmentation method based on the Adaptive Concentric Zone Model achieves favorable segmentation performance on both types of data, indicating strong adaptability and robustness across different data sources and road scenarios. Further comparison shows that the proposed method performs better on the self-collected RUBY PLUS point cloud than on the SemanticKITTI dataset. The primary reason is that only the three-dimensional spatial coordinates of the point clouds are used in the SemanticKITTI experiments. Under this condition, the method mainly relies on Adaptive Concentric Zone partitioning, weighted PCA-based local plane fitting, and geometric consistency constraints to extract candidate ground points. In contrast, the point clouds acquired using the RUBY PLUS 128-channel LiDAR contain both three-dimensional spatial coordinates and reflectance intensity information, allowing the Reflectance Intensity Consistency Constraint in the fine segmentation stage to be fully exploited.

In the experiments on the self-collected RUBY PLUS point clouds, the Reflectance Intensity Consistency Constraint further distinguishes high-confidence ground points from uncertain points among the candidate ground points obtained by coarse segmentation. The uncertain points are then re-evaluated using the local height stability of high-confidence neighborhoods, thereby effectively reducing the misclassification of non-ground structures and the missed detection of true ground points. Therefore, the more complete point attributes available in the self-collected data enhance the discriminative capability of the fine segmentation stage, enabling the proposed method to achieve higher Precision, Recall, and F1-score. Overall, the experiments on the public dataset verify the effectiveness of the proposed method when only spatial geometric information is available, whereas the experiments on the self-collected RUBY PLUS data further demonstrate that, when both spatial coordinates and reflectance intensity information are provided, the proposed two-stage segmentation framework can more fully exploit the complementary constraints of intensity consistency and local height stability, thereby achieving more stable, complete, and accurate ground segmentation.

### 4.3 Ablation studies

To verify the effectiveness of the key modules in the proposed method, a module-level ablation study was conducted. The proposed framework consists of three principal components: Adaptive Concentric Zone partitioning, first-stage coarse segmentation, and second-stage fine segmentation. Accordingly, a progressive ablation scheme was constructed by incrementally incorporating the key modules into a baseline model. The baseline model employs uniform polar partitioning, standard PCA-based plane fitting, and a point-to-plane distance criterion. On this basis, the following components are successively introduced: Adaptive Concentric Zone partitioning; the geometric coarse segmentation module, comprising the lowest-height seed constraint and height-decay weighted PCA; and the fine segmentation module, comprising the Reflectance Intensity Consistency Constraint and local height stability refinement. The contributions of the individual modules to ground segmentation performance are analyzed by comparing Precision, Recall, and F1-score under different module combinations.

Fig. 6 presents the visual segmentation results of the different ablation models on the self-collected RUBY PLUS LiDAR point cloud. Green points denote correctly segmented ground points; gray points denote correctly segmented non-ground points; red points denote non-ground points misclassified as ground; and blue points denote ground points misclassified as non-ground. As shown in the figure, the baseline model A0 produces numerous red and blue points, particularly near road boundaries, local obstacles, and sparse point cloud regions. This indicates substantial false detection of non-ground points and missed detection of true ground points, demonstrating that uniform partitioning, standard PCA-based plane fitting, and a distance threshold alone cannot robustly accommodate complex road environments. After introducing Adaptive Concentric Zone partitioning, A1 exhibits improved ground continuity compared with A0, and fewer blue false-negative points are observed over several major road regions. This result indicates that adaptive partitioning can alleviate the local modeling imbalance caused by the range-dependent density distribution of vehicle-mounted LiDAR point clouds. Nevertheless, A1 still contains numerous red false-positive points, suggesting that improving the partitioning strategy alone is insufficient to suppress interference from local non-ground structures during plane estimation. After further incorporating the geometric coarse segmentation module, A2 produces a more complete distribution of green ground points, while both red false-positive points and blue false-negative points are substantially reduced. This demonstrates that the lowest-height seed constraint, height-decay weighted PCA, and geometric consistency constraints effectively improve the robustness of local ground plane estimation and reduce the influence of non-ground points associated with vehicles, vegetation, and roadside structures. In the complete model A3, the green point coverage shows the highest agreement with the ground-truth ground region, and the red and blue error points are further reduced. In particular, A3 exhibits greater classification stability near road boundaries and in locally mixed-structure regions. These results indicate that the intensity-guided fine segmentation module can further distinguish high-confidence ground points from uncertain points by exploiting reflectance intensity consistency and local height stability based on the geometric coarse segmentation results, thereby reducing the misclassification of non-ground points and recovering some missed ground points.

Overall, from A0 to A3, both false-positive and false-negative points are progressively reduced, while the continuity and completeness of the segmented ground region are gradually improved. These results verify the effectiveness of Adaptive Concentric Zone partitioning, geometric coarse segmentation, and intensity-guided fine segmentation in improving ground segmentation performance for real-world road point clouds.

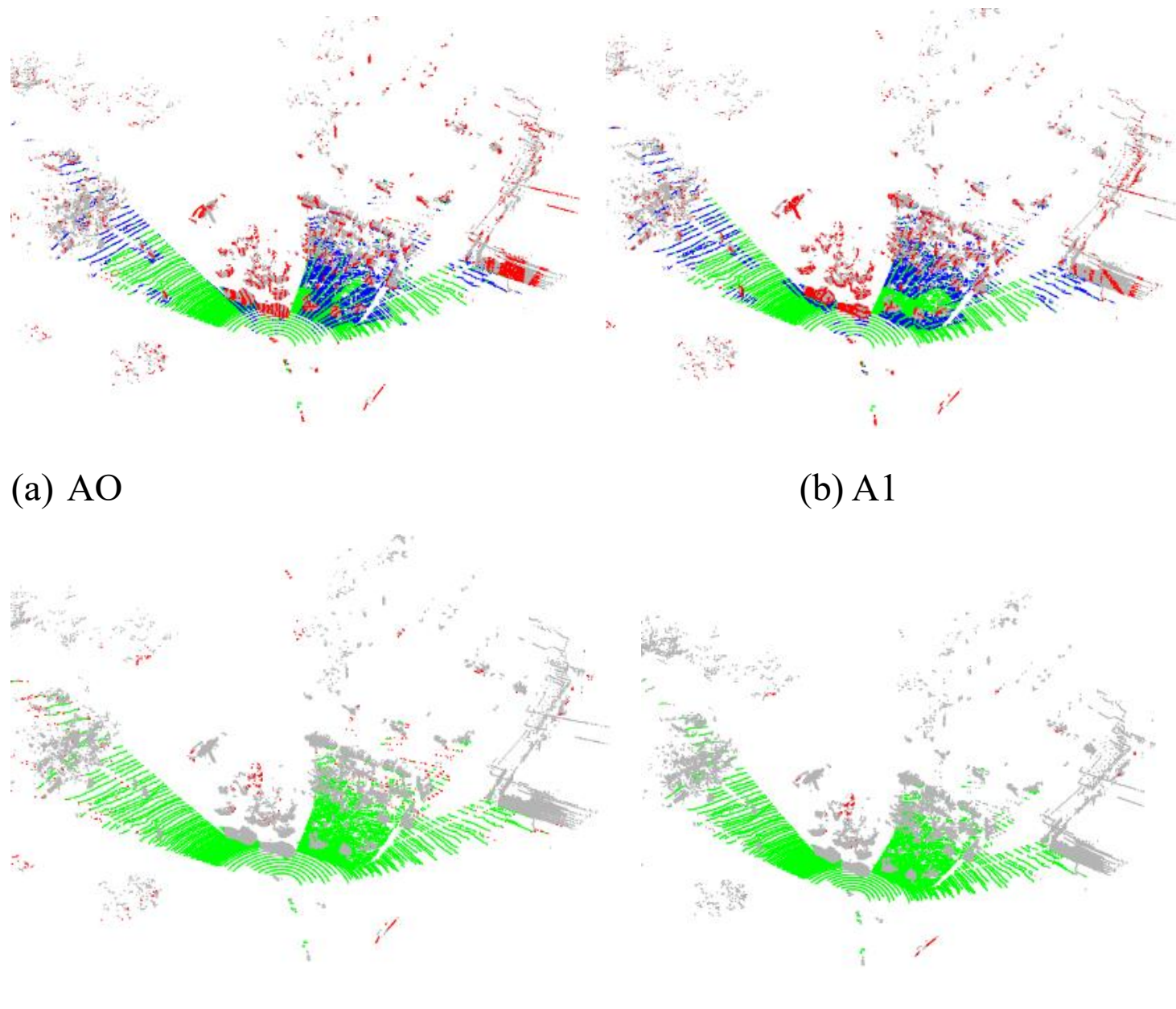

Fig. 6 Visual comparison of the segmentation results obtained by different ablation models

As shown in Table 5, the baseline model A0 employs only uniform polar partitioning and a basic plane classification strategy,

achieving a Precision of 10.93%, a Recall of 65.90%, and an F1-score of 18.75%. Its overall performance is therefore relatively poor. This result indicates that, in real-world road point clouds, basic geometric discrimination alone cannot effectively handle range-dependent point density, local ground undulations, and interference from non-ground structures, leading to substantial false detections and missed detections. After introducing Adaptive Concentric Zone partitioning, A1 achieves a certain improvement over A0. Specifically, Precision increases from 10.93% to 21.96%, Recall from 65.90% to 72.42%, and F1-score from 18.75% to 33.70%. These results demonstrate that Adaptive Concentric Zone partitioning can improve the local point distribution across different ranges and enhance the stability of local modeling in far-range sparse regions. Nevertheless, the partitioning strategy alone remains insufficient to effectively suppress the influence of non-ground structures on ground plane estimation.

After further incorporating the geometric coarse segmentation module, A2 achieves a Precision of 97.11%, a Recall of 100.00%, and an F1-score of 98.53%, representing improvements of 75.15, 27.58, and 64.83 percentage points over A1, respectively. This constitutes the most substantial performance gain among all ablation stages. The result demonstrates that the geometric coarse segmentation module, comprising the lowest-height seed constraint, height-decay weighted PCA, Planarity Constraint, and Normal Angle Constraint, is the principal contributor to the improvement in ground segmentation performance. This module effectively reduces the influence of non-ground points associated with vehicles, vegetation, and roadside structures on local plane fitting, while improving the accuracy and completeness of candidate ground point extraction.

After the intensity-guided fine segmentation module is further added to A2, the complete model A3 achieves a Precision of 98.72%, a Recall of 100.00%, and an F1-score of 99.36%, yielding the best overall performance. Compared with A2, A3 maintains Recall at 100.00%, while Precision and F1-score increase by 1.61 and 0.83 percentage points, respectively. This indicates that the Reflectance Intensity Consistency Constraint and local height stability refinement can further reduce the misclassification of non-ground points without sacrificing ground point recall. Overall, the progressive improvement from A0 to A3 confirms the effectiveness of Adaptive Concentric Zone partitioning, geometric coarse segmentation, and intensity-guided fine segmentation. Among these components, the geometric coarse segmentation module contributes the largest performance gain. In contrast, the intensity-guided fine segmentation module further improves false-positive suppression, enabling the proposed method to achieve higher segmentation accuracy and robustness on real-world RUBY PLUS LiDAR point clouds.

Table 5 Quantitative results of the ablation study

| Method | Adaptive Concentric Zone Module | Coarse Segmentation Module | Fine Segmentation Module | Precision (%) | Recall (%) | F1 (%) |
|---|---|---|---|---|---|---|
| A0 | × | × | × | 10.93 | 65.90 | 18.75 |
| A1 | ✓ | × | × | 21.96 | 72.42 | 33.70 |
| A2 | ✓ | ✓ | × | 97.11 | 100.00 | 98.53 |
| A3(ours) | ✓ | ✓ | ✓ | 98.72 | 100.00 | 99.36 |

## 4.4 Computational Complexity Analysis

To evaluate the computational cost of the proposed two-stage ground segmentation method based on the Adaptive Concentric Zone Model, the time and space complexities of each module are analyzed according to the processing pipeline described above. Let the input point cloud be denoted as

$$P=\{(p_i,I_i)\}_{i=1}^{N} \tag{68}$$

where N denotes the total number of points. After Adaptive Concentric Zone partitioning, suppose that K valid rings are obtained and that the k-th ring contains $s_k$ sectors. The total number of local zones is therefore $\sum_{k=1}^{K} s_k$. Let $\beta_{k,l}$ denote the local zone corresponding to the l-th sector of the k -th ring, and let $n_{k,l}$ denote the number of points contained in $\beta_{k,l}$, satisfying

$$\sum_{k=1}^{K}\sum_{l=1}^{s_k} n_{k,l} = N \tag{69}$$

Let G denote the candidate ground point set obtained by the first-stage coarse segmentation, with cardinality $|G|$. In the second stage, let $G^{\mu}$ denote the uncertain point set identified by the reflectance intensity consistency criterion, with cardinality $|G^{\mu}|$. For each candidate point $p_i \in G$, let $q_j$ denote the number of neighboring points within its local neighborhood in the high-confidence ground point set. The total time complexity of the proposed method is given by:

$$T_{total} = \underbrace{\left(c_{p1}N + c_{p2}N\log N + c_{p2}N\right)}_{Adaptive\ Concentric\ Zone\ Module} + \underbrace{\sum_{k=1}^{K}\sum_{l=1}^{s_k}\left(c_{f1}n_{k,l} + c_{f2}\right)}_{Coarse\ Segmentation\ Module} + \underbrace{\left(c_{r1}|G|\log|G| + c_{r2}\sum_{p_i\in G} m_i + c_{r3}|G^{\mu}|\log|G| + c_{r4}\sum_{p_j\in G^{\mu}} q_j\right) + c_{m1}|G|\log|G|}_{Fine\ Segmentation\ Module} \tag{70}$$

Where $c_{n1}, c_{n2}, c_{n3}, c_{f1}, c_{f2}, c_{r1}, c_{r2}, c_{r3}, c_{r4}, c_{m1}$ denote constant-order implementation costs associated with the individual modules. These constants do not affect the asymptotic growth of the computational complexity with respect to the number of points. As shown in Eq. (70), the primary computational overhead of the proposed method arises from four components: polar coordinate transformation and quantile-based ring construction, local weighted PCA-based plane fitting, second-stage neighborhood refinement, and MAD-based robust secondary removal of high-elevation points. Since $|G| \le N$, and the average size of a local neighborhood can generally be regarded as bounded under a fixed neighborhood radius R, it follows that

$$T_{\text{total}} = O\left(N\log N\right) \tag{71}$$

The proposed method maintains favorable computational scalability while ensuring segmentation accuracy and robustness. The computational complexity of each module is analyzed separately below.

#### 4.4.1 Adaptive Concentric Zone Model

First, the horizontal radial distance $r_i$ and azimuth angle $\theta_i$ are computed for each point in the point cloud P. Since this process involves only a constant number of arithmetic operations per point, its time complexity is $O\left(N\right)$. Subsequently, to construct quantile-based rings with approximately equal numbers of points, quantiles must be computed from all radial distance samples $\{r_i\}_{i=1}^{N}$. When sorting is used, this step requires $O\left(N\log N\right)$ time. After the ring boundaries are determined, each point is assigned a ring index and a sector index according to $\left(r_i, \theta_i\right)$, and is then allocated to the corresponding local zone $B_{k.l}$. Because this process requires only a constant number of indexing and interval-comparison operations for each point, its time complexity is $O\left(N\right)$. Therefore, the total time complexity of the Adaptive Concentric Zone partitioning module is:

$$O\left(N\right) + O\left(N\log N\right) + O\left(N\right) = O\left(N\log N\right) \tag{72}$$

From the perspective of the dominant term, the principal computational cost of the Adaptive Concentric Zone partitioning stage arises from the sorting operation required for quantile-based ring construction, whereas the polar coordinate transformation and point-to-zone assignment incur only linear costs. This stage corresponds to the first term in Eq. (72).

#### 4.4.2 Coarse Segmentation

For each local zone $B_{k,l}$, the following operations are performed sequentially: lowest-height point search, construction of the fitting-point subset under the seed-band constraint, computation of height-decay weights, construction of the weighted centroid and weighted covariance matrix, and candidate ground point classification based on planarity, normal angle, and point-to-plane distance criteria. For any local zone $B_{k,l}$, these operations require only a constant number of scans or statistical calculations over its $n_{k,l}$ points, resulting in a time complexity of $O(n_{k,l})$. In addition, since the covariance matrix has a fixed size of $3\times3$, its eigenvalue decomposition can be regarded as a constant-time operation, $O(1)$, and therefore does not alter the asymptotic complexity of the local zone. Accordingly, the processing cost of the $(k,l)$-th local zone can be expressed as:

$$O\left(n_{k,l}\right) + O\left(1\right) = O\left(n_{k,l}\right) \tag{73}$$

Summing over all local zones gives:

$$\sum_{k=1}^{K}\sum_{l=1}^{s_k} O\left(n_{k,l}\right) = O\left(\sum_{k=1}^{K}\sum_{l=1}^{s_k} n_{k,l}\right) = O\left(N\right) \tag{74}$$

Therefore, although this stage introduces the lowest-height seed constraint and the height-decay weighting strategy, the overall complexity of local plane fitting and coarse segmentation remains linear. This indicates that the proposed local geometric modeling module does not become a major computational bottleneck as the number of points increases. This component corresponds to the second term in Eq. (74).

#### 4.4.3 Fine Segmentation

After the candidate ground point set G is obtained in the first stage, the second stage first performs reflectance intensity consistency analysis within a radius-R neighborhood to divide the candidate points into high-confidence ground points and uncertain points. Subsequently, the uncertain point set $G^u$ is refined through a restoration procedure based on the local height standard deviation within high-confidence ground neighborhoods. The main computational cost of this stage arises from neighborhood queries. Without a spatial indexing structure, directly performing a global search for each point would result in a worst-case complexity of $O(|G|^2)$. In practical implementation, however, a kd-tree or an equivalent two-dimensional spatial index can first be constructed over the candidate ground point set G, with a construction complexity of $O(|G|\log|G|)$.

On this basis, a radius query is performed for each candidate ground point $p_i \in G$, followed by computation of the mean reflectance intensity difference within its neighborhood. The computational cost for each query can be expressed as $O(\log|G|+m_i)$. Therefore, the total complexity of reflectance intensity consistency classification is given by:

$$O\left(\sum_{p_i\in G}\left(\log|G|+m_i\right)\right)=O\left(|G|\log|G|\right)+O\left(\sum_{p_i\in G}m_i\right) \quad (75)$$

Similarly, for each uncertain point $p_j \in G^u$, a neighborhood query is performed within the high-confidence ground point set, followed by computation of the local height standard deviation. The computational cost of this operation is $O(\log|G|+q_j)$. Therefore, the total complexity of uncertain-point refinement is given by:

$$O\left(\sum_{p_j\in G^u}\left(\log|G|+q_j\right)\right)=O\left(|G^u|\log|G|\right)+O\left(\sum_{p_j\in G^u}q_j\right) \quad (76)$$

Combining the two components above, the total complexity of this module corresponds to the third term in Eq. (76). Furthermore, in vehicle-mounted LiDAR scenarios, when the neighborhood radius R is fixed and the local point distribution becomes relatively balanced after Adaptive Concentric Zone partitioning, the average neighborhood sizes $1/|G|\sum_{p_i\in G}m_i$ and $1/|G^u|\sum_{p_j\in G^u}q_j$ can generally be regarded as bounded constants. Therefore, the complexity of this module can be approximately simplified to $O\left(|G|\log|G|\right)$.

Since $|G|\le N$, the asymptotic upper bound of this stage is $O(N\log N)$. This indicates that although the second stage introduces neighborhood consistency verification and local height constraints, its computational complexity remains manageable when an efficient spatial indexing structure is employed.

After intensity consistency refinement, robust statistical analysis is further performed on the height components of the final ground point set G, including the computation of the median and median absolute deviation MAD, to construct an adaptive upper height threshold for removing misclassified high-elevation points. When a standard sorting-based implementation is adopted, the computational complexities of both the median and MAD calculations are $O\left(|G|\log|G|\right)$. Therefore, the computational complexity of this module is $O\left(|G|\log|G|\right)$.

Based on the preceding analysis, the total computational complexity of the proposed method is given by:

$$T_{total}=O\left(N\log N\right)+O\left(N\right)+O\left(|G|\log|G|\right)+O\left(|G|\log \quad (77)$$

Since $|G|\le N$, the complexity can be simplified to:

$$T_{total}=O\left(N\log N\right) \quad (78)$$

In summary, the proposed two-stage ground segmentation method based on the Adaptive Concentric Zone Model has an overall time complexity of $O(N\log N)$ and achieves relatively fast execution compared with other methods, as shown in Table 6. The dominant computational costs arise primarily from quantile-based ring partitioning, spatial index construction for neighborhood searches in the second stage, and sorting operations in robust height statistics. In contrast, weighted PCA-based plane fitting and candidate ground point extraction within local zones retain linear complexity, $O(N)$. Therefore, the proposed method exhibits favorable engineering scalability and strong potential for real-time deployment in vehicle-mounted LiDAR point cloud applications.

Table 6 Speed comparison of ground segmentation methods on Sequence 05 of the SemanticKITTI dataset using an AMD Ryzen 7 4800H with Radeon Graphics CPU (unit: Hz)

| Method | Speed | Method | Speed |
|---|---|---|---|
| DBSCAN | 32.77 | FugSeg | 41.69 |
| TRAVEL | 15.32 | PatchWork | 43.97 |
| GroundGrid | 7.41 | PatchWork++ | 54.58 |
| Ours | | 30.91 | |

# 5 Conclusion

To address the challenges posed by range-dependent point density, local ground undulations, and interference from non-ground structures in local plane estimation under complex road conditions, this study proposes ACZ-GSeg, a two-stage ground segmentation method based on an Adaptive Concentric Zone Model. An Adaptive Concentric Zone Model is first constructed, upon which the proposed two-stage ground segmentation framework is developed. Experimental results on the public SemanticKITTI dataset and self-collected RUBY PLUS LiDAR point cloud demonstrate that ACZ-GSeg effectively improves the completeness, accuracy, and robustness of ground segmentation. The ablation study further verifies the effectiveness of Adaptive Concentric Zone partitioning, the geometric coarse segmentation module, and the intensity-guided fine segmentation module. These results indicate that the proposed method can provide reliable ground priors for subsequent tasks, including object detection, clustering, tracking, and traversable-area analysis.

Future work will focus on the following aspects. (1) ACZ-GSeg contains multiple parameters, which may increase the tuning cost when the method is applied to different sensors and scenarios. Future research will therefore investigate adaptive parameter estimation and parameter simplification strategies to reduce dependence on manual tuning. (2) Further acceleration remains necessary for large-scale point clouds acquired by high-channel-count LiDAR sensors and for high-frame-rate online perception scenarios. Parallel zone processing, efficient neighborhood search, and GPU acceleration will be explored to reduce computational overhead and improve the deployment efficiency of ACZ-GSeg in real-time perception systems for ground-based mobile platforms.

# Declaration of competing interest

The authors declare that they have no known competing financial interests or personal relationships that could have appeared to influence the work reported in this paper.